\documentclass{article}

\usepackage[utf8]{inputenc} 
\usepackage[T1]{fontenc}    
\usepackage{hyperref}       
\usepackage{url}            
\usepackage{booktabs}       
\usepackage{amsfonts}       
\usepackage{nicefrac}       
\usepackage{microtype}      
\usepackage{graphicx}
\usepackage[small]{titlesec}
\usepackage{verbatim}
\usepackage{amsmath}
\usepackage{cleveref}

\usepackage{rotating}

\usepackage[T1]{fontenc}
\usepackage{lmodern}
\usepackage{authblk}

\usepackage{siunitx,caption}

\usepackage{lineno}

\usepackage[verbose=true,letterpaper]{geometry}
 \newgeometry{
    textheight=9in,
    textwidth=6.27in,
    top=1in,
    headheight=12pt,
    headsep=25pt,
    footskip=30pt
  }

\title{\textbf{Analyzing Neuroimaging Data Through \\ Recurrent Deep Learning Models}}

\author[1,2,3]{\small Armin W. Thomas}
\author[2,3 *]{\small Hauke R. Heekeren}
\author[1,4,5 *]{\small Klaus-Robert Müller}
\author[6 *]{\small Wojciech Samek}
\affil[1]{Machine Learning Group, Technische Universität Berlin, Marchstr. 23, Berlin 10587, Germany}
\affil[2]{Center for Cognitive Neuroscience Berlin, Freie Universität Berlin, Habelschwerdter Allee 45, Berlin 14195, Germany}
\affil[3]{Department of Education \& Psychology, Freie Universität Berlin, Habelschwerdter Allee 45, Berlin 14195, Germany}
\affil[4]{Department of Brain \& Cognitive Engineering, Korea University, Anam-dong 5ga, Seongbuk-gu, Seoul 136-713, South Korea}
\affil[5]{Max Planck Institute for Informatics, Stuhlsatzenhausweg, Saarbrücken 66123, Germany}
\affil[6]{Machine Learning Group, Fraunhofer Heinrich Hertz Institute, Einsteinufer 37, Berlin 10587, Germany}
\affil[*]{Corresponding authors: hauke.heekeren@fu-berlin.de, klaus-robert.mueller@tu-berlin.de,  wojciech.samek@hhi.fraunhofer.de}
\date{}

\begin{document}
\maketitle

\noindent\makebox[\linewidth]{\rule{\linewidth}{0.4pt}}
\section*{Abstract}
The application of deep learning (DL) models to neuroimaging data poses several challenges, due to the high dimensionality, low sample size and complex temporo-spatial dependency structure of these datasets. Even further, DL models act as as {\em black-box} models, impeding insight into the association of cognitive state and brain activity. To approach these challenges, we introduce the DeepLight framework, which utilizes long short-term memory (LSTM) based DL models to analyze {\em whole-brain} functional Magnetic Resonance Imaging (fMRI) data. To decode a cognitive state (e.g., seeing the image of a house), DeepLight separates the fMRI volume into a sequence of axial brain slices, which is then sequentially processed by an LSTM. To maintain interpretability, DeepLight adapts the layer-wise relevance propagation (LRP) technique. Thereby, decomposing its decoding decision into the contributions of the single input voxels to this decision. Importantly, the decomposition is performed on the level of single fMRI volumes, enabling DeepLight to study the associations between cognitive state and brain activity on several levels of data granularity, from the level of the group down to the level of single time points. To demonstrate the versatility of DeepLight, we apply it to a large fMRI dataset of the Human Connectome Project. We show that DeepLight outperforms conventional approaches of uni- and multivariate fMRI analysis in decoding the cognitive states and in identifying the physiologically appropriate brain regions associated with these states. We further demonstrate DeepLight's ability to study the fine-grained temporo-spatial variability of brain activity over sequences of single fMRI samples.
\noindent\makebox[\linewidth]{\rule{\linewidth}{0.4pt}}
\textbf{Keywords}: Decoding, fMRI, whole-brain, deep learning, LSTM, interpretability.\\

\section{Introduction}
\label{sec:introduction}
Neuroimaging research has recently started collecting large corpora of experimental functional Magnetic Resonance Imaging (fMRI) data, often comprising many hundred individuals \cite{van2013wu, poldrack2013toward}. By collecting these datasets, researchers want to gain insights into the associations between the cognitive states of an individual (e.g., while viewing images or performing a specific task) and the underlying brain activity, while also studying the variability of these associations across the population \cite{van2013wu}. 

At first sight, the analysis of neuroimaging data thereby seems ideally suited for the application of deep learning (DL) \cite{lecun2015deep, goodfellow2016deep} methods, due to the availability of large and structured datasets. Generally, DL can be described as a class of representation-learning methods, with multiple levels of abstraction. At each level, the representation of the input data is transformed by a simple, but non-linear function. The resulting hierarchical structure of non-linear transforms enables DL methods to learn complex functions. It also enables them to identify intricate signals in noisy data, by projecting the input data into a higher-level representation, in which those aspects of the input data that are irrelevant to identify an analysis target are suppressed and those that are relevant are amplified. With this higher-level perspective, DL methods can associate a target variable with variable patterns in the input data. Importantly, DL methods can autonomously learn these projections from the data and therefore do not require a thorough prior understanding of the mapping between input data and analysis target (for a detailed discussion, see \cite{lecun2015deep}). For these reasons, DL methods seem ideally suited for the analysis of neuroimaging data, where intricate, highly variable patterns of brain activity are hidden in large, high-dimensional datasets and the mapping between cognitive state and brain activity is often unknown. 

While researchers have started exploring the application of DL models to neuroimaging data (e.g., \cite{plis2014deep, mensch2018extracting, sarraf2016classification, nie20163d, suk2014hierarchical, petrov2018deep, yousefnezhad2018anatomical}), two major challenges have so far prevented broad DL usage: (1) Neuroimaging data are high dimensional, while containing comparably few samples. For example, a typical fMRI dataset comprises up to a few hundred samples per subject and recently up to several hundred subjects \cite{van2013wu}, while each sample contains several hundred thousand dimensions (i.e., voxels). In such analysis settings, DL models (as well as more traditional machine learning approaches) are likely to suffer from overfitting (by too closely capturing those dynamics that are specific to the training data so that their predictive performance does not generalize well to new data). (2) DL models have often been considered as non-linear {\em black box models}, disguising the relationship between input data and decoding decision. Thereby, impeding insight into (and interpretation of) the association between cognitive state and brain activity.

To approach these challenges, we propose the DeepLight framework, which defines a method to utilize long short-term memory (LSTM) based DL architectures \cite{hochreiter1997long, donahue2015long} to analyze whole-brain neuroimaging data. In DeepLight, each whole-brain volume is sliced into a sequence of axial images. To decode an underlying cognitive state, the resulting sequence of images is processed by a combination of convolutional and recurrent DL elements. Thereby, DeepLight successfully copes with the high dimensionality of neuroimaging data, while modeling the full spatial dependency structure of whole-brain activity (within and across axial brain slices). Conceptually, DeepLight builds upon the searchlight approach. Instead of moving a small searchlight beam around in space, DeepLight explores brain activity more in-depth, by looking through the full sequence of axial brain slices, before making a decoding decision. To subsequently relate brain activity and cognitive state, DeepLight applies the layer-wise relevance propagation (LRP) \cite{BachPLOS15,LapJMLR16} method to its decoding decisions. Thereby, decomposing these decisions into the contributions of the single input voxels to each decision. Importantly, the LRP analysis is performed on the level of a single input samples, enabling an analysis on several levels of data granularity, from the level of the group down to the level of single subjects, trials and time points. These characteristics make DeepLight ideally suited to study the fine-grained temporo-spatial distribution of brain activity underlying sequences of single fMRI samples.

Here, we will demonstrate the versatility of DeepLight, by applying it to an openly available fMRI dataset of the Human Connectome Project \cite{van2013wu}. In particular, to the data of an N-back task, in which 100 subjects viewed images of either body parts, faces, places or tools in two separate fMRI experiment runs (for an overview, see Supplementary Fig.\ S1 and Section \ref{sec:experiment_paradigm_details}). Subsequently, we will evaluate the performance of DeepLight in decoding the four underlying cognitive states (resulting from viewing an image of either of the four stimulus classes) from the fMRI data and identifying the brain regions associated with these states. To this end, we will compare the performance of DeepLight to three representative conventional approaches to the uni- and multivariate analysis of neuroimaging data. In particular, the General Linear Model (GLM) \cite{friston1994statistical}, searchlight analysis \cite{kriegeskorte2006information} and whole-brain Least Absolute Shrinkage Logistic Regression (whole-brain Lasso) \cite{wager2013fmri, grosenick2013interpretable}. Importantly, these approaches differ in the number of voxels they include in their analyses. While the GLM analyses the data of single voxels independent of one another (univariate), the searchlight analysis utilizes the data of clusters of multiple voxels (multivariate) and the whole-brain lasso utilizes the data of all voxels in the brain (whole-brain). In this comparison, we find that DeepLight (1) decodes the cognitive states underlying the fMRI data more accurately than these other approaches, (2) improves its decoding performance better with growing datasets, (3) accurately identifies the physiologically appropriate associations between cognitive states and brain activity and (4) identifies these associations on multiple levels of data granularity (namely, the level of the group, subject, trial and time point). We also demonstrate DeepLight's ability to study the temporo-spatial distribution of brain activity over a sequence of single fMRI samples.

\section{Methods}

\subsection{Experiment paradigm}
\label{sec:experiment_paradigm_details}
100 participants performed a version of the N-back task in two separate fMRI runs (for an overview, see Supplementary Fig.\ S1 as well as \cite{barch2013function}). Each of the two runs (260s each) consisted of eight task blocks (25s each) and four fixation blocks (15s each). Within each run, the four different stimulus types (body, face, place and tool) were presented in separate blocks. Half of the task blocks used a 2-back working memory task (participants were asked to respond "target" when the current stimulus was the same as the stimulus 2 back) and the other half a 0-back working memory task (a target cue was presented at the beginning of each block and the participants were asked to respond "target" whenever the target cue was presented in the block). Each task block consisted of 10 trials (2.5s each). In each trial, a stimulus was presented for 2s followed by a 500 ms interstimulus interval (ISI). We were not interested in identifying any effect of the N-back task condition on the evoked brain activity and therefore pooled the data of both N-back conditions.

\subsection{FMRI data acquisition \& preprocessing}
\label{sec:fmri_data_acquisition_details}
Functional MRI data of 100 unrelated participants for this experiment were provided in a preprocessed format by the Human Connectome Project, WU Minn Consortium (Principal Investigators: David Van Essen and Kamil Ugurbil; 1U54MH091657) funded by the 16 NIH Institutes and Centers that support the NIH Blueprint for Neuroscience Research; and by the McDonnell Center for Systems Neuroscience at Washington University. Whole-brain EPI acquisitions were acquired with a 32 channel head coil on a modified 3T Siemens Skyra with TR=720 ms, TE=33.1 ms, flip angle=52 deg, BW=2290 Hz/Px, in-plane FOV=$208 \times 180 mm$, 72 slices, 2.0 mm isotropic voxels with a multi-band acceleration factor of 8. Two runs were acquired, one with a right-to-left and the other with a left-to-right phase encoding (for further methodological details on fMRI data acquisition, see \cite{uugurbil2013pushing}). 

\label{sec:fmri_preprocessing_details}
The Human Connectome Project preprocessing pipeline for functional MRI data ("fMRIVolume") \cite{glasser2013minimal} includes the following steps: gradient unwarping, motion correction, fieldmap-based EPI distortion correction, brain-boundary based registration of EPI to structural T1-weighted scan, non-linear registration into MNI152 space, and grand-mean intensity normalization (for further details, see \cite{uugurbil2013pushing, glasser2013minimal}).
In addition to the minimal preprocessing of the fMRI data that was performed by the Human Connectome Project, we applied the following preprocessing steps to the data for all decoding analyses: volume-based smoothing of the fMRI sequences with a 3mm Gaussian kernel, linear detrending and standardization of the single voxel signal time-series (resulting in a zero-centered voxel time-series with unit variance) and temporal filtering of the single voxel time-series with a butterworth highpass filter and a cutoff of 128s, as implemented in Nilearn 0.4.1  \cite{abraham2014machine}.
In line with previous work \cite{jang2017task}, we further applied an outer brain mask to each fMRI volume. We first identified those voxels whose activity was larger than 5\% of the maximum voxel signal within the fMRI volume and then only kept those voxels for further analysis that were positioned between the first and last voxel to fulfill this property in the three spatial dimensions of any functional brain volume of our dataset. This resulted in a brain mask spanning $74 \times 92 \times 81$ voxels $(X \times Y \times Z)$.

All of our preprocessing was performed by the use of the Nilearn 0.4.1 Python library \cite{abraham2014machine}. Importantly, we did not exclude any TR of an experiment block of the four stimulus classes from the decoding analyses. However, we removed all fixation blocks from the decoding analyses. Lastly, we split the fMRI data of the 100 subjects contained in the dataset into two distinct training and test datasets (each containing the data of 70 and 30 randomly assigned subjects). All analyses presented throughout the following solely include the data of the 30 subjects contained in the held-out test dataset (if not stated otherwise).

\subsection{Baseline methods}
\subsubsection{General linear model}
\label{sec:GLM_details}
The General Linear Model (GLM) \cite{friston1994statistical} represents a univariate brain encoding model \cite{naselaris2011encoding, kriegeskorte2018interpreting}. Its goal is to identify an association between cognitive state and brain activity, by predicting the time series signal of a voxel from a set of experiment predictors:

\begin{equation}
	Y = X\beta + \epsilon
\end{equation}

Here, $Y$ presents a $T \times N$ dimensional matrix containing the multivariate time series data of $N$ voxels and $T$ time points. $X$ represents the design matrix, which is composed of $T \times P$ data points, where each column represents one of $P$ predictors. Typically, each predictor represents a variable that is manipulated during the experiment (e.g., the presentation times of stimuli of one of the four stimulus classes). $\beta$ represents a $P \times N$ dimensional matrix of regression coefficients. To mimic the blood-oxygen-level dependent (BOLD) response measured by the fMRI, each predictor is first convolved with a hemodynamic response function (HRF), before fitting the $\beta$ coefficients to the data. After fitting, these coefficients indicate the estimated contribution of each predictor to the time series signal of each of the $N$ voxels. $\epsilon$ represents a $T \times N$ dimensional matrix of error terms. Importantly, the GLM analyzes the time series signal of each voxel independently and thereby includes a separate set of regression coefficients for each voxel in the brain.

\subsubsection{Searchlight analysis}
\label{sec:searchlight_details}
The searchlight analysis is a multivariate brain decoding model \cite{kriegeskorte2006information}. Its goal is to identify an association between cognitive state and brain activity, by probing the ability of a statistical classifier to decode the cognitive state from the activity pattern of a small clusters of voxels. To this end, the entire brain is scanned with a sphere of a given radius (the searchlight) and the performance of the classifier in decoding the cognitive states is evaluated at each location, resulting in a map of decoding accuracies. Here, we used a searchlight radius of 5.6mm and a linear-kernel Support Vector Machine (SVM) classifier (if not reported otherwise).

Given a training dataset of $T$ data points ${[y_{t}, x_{t}]}_{t=1}^{T}$, where $x_{t}$ represents the input pattern and $y_{t}$ the label of data point $t$, the SVM \cite{cortes1995support} is defined as follows:  

\begin{equation}
	\hat{y}(x) = \mathrm{sign}[\sum_{t=1}^{T} \alpha_{t} y_{t} \gamma(x, x_{t}) + b]
\end{equation}

Here, $\alpha_{t}$ and $b$ are positive constants, whereas $\gamma(x, x_{t})$ represents the kernel function. We used a linear kernel function, as implemented in Nilearn 0.4.1 \cite{abraham2014machine}. We then defined the decoding accuracy of the searchlight analysis as the maximum decoding accuracy that was achieved at any searchlight location in the brain.

\subsubsection{Whole-brain Least Absolute Shrinkage Logistic Regression}
\label{sec:wholebrain_lasso_details}
The whole-brain Least Absolute Shrinkage Logistic Regression (or whole-brain lasso) represents a whole-brain decoding model. It identifies an association between cognitive state and brain activity, by probing the ability of a logistic model to decode the cognitive state from whole-brain activity (with one logistic coefficient $\beta_{i}$ per voxel $i$ in the brain). To reduce the risk of overfitting, resulting from the large number of model coefficients, the whole-brain lasso applies Least Absolute Shrinkage regularization to the likelihood function of the logistic model \cite{tikhonov1943stability, tibshirani1996regression}. Thereby, forcing the logistic model to perform automatic variable selection during parameter estimation, resulting in sparse coefficient estimates (i.e., by forcing many coefficient estimates to be exactly 0). In particular, the optimization problem of the whole-brain lasso can be defined as follows (again, $N$ denotes the number of voxels in the brain, $T$ the number of fMRI sampling time points and $[y_{t}, x_{t}]_{t=1}^{T}$ the set of class labels and voxel values of each fMRI sample):

\begin{equation}
    \label{eq:4}
	\min_{\beta}\Bigg\{{\sum_{t=1}^{T}\bigg[{y_{t} \log\sigma(\beta^{T}x_{t}) + (1 -  y_{t})\log(1 - \sigma(\beta^{T}x_{t}))}}\bigg] + \lambda \sum_{i=1}^{N}|\beta_{i}| \Bigg\}
\end{equation}

Here, $\lambda$ represents the strength of the regularization term (with larger values indicating stronger regularization), whereas $\sigma$ represents the logistic model:

\begin{equation}
    \label{eq:logit}
    \sigma(z) = \frac{1}{1+e^{-z}} 
\end{equation}

The resulting set of coefficient estimates $\beta$, indicates the contribution of each voxel $i$ to the decoding decision of the logistic regression model. Over the recent years, the whole-brain lasso, as well as closely related decoding approaches (e.g., \cite{ryali2010sparse, gramfort2013identifying, mcintosh2004partial}), have found widespread application throughout the neuroscience literature (e.g., \cite{chang2015sensitive, wager2013fmri}).

\subsection{DeepLight framework}
\label{sec:DeepLight_details}

\begin{figure*}[!ht]
    \centering
    \includegraphics[width=\linewidth]{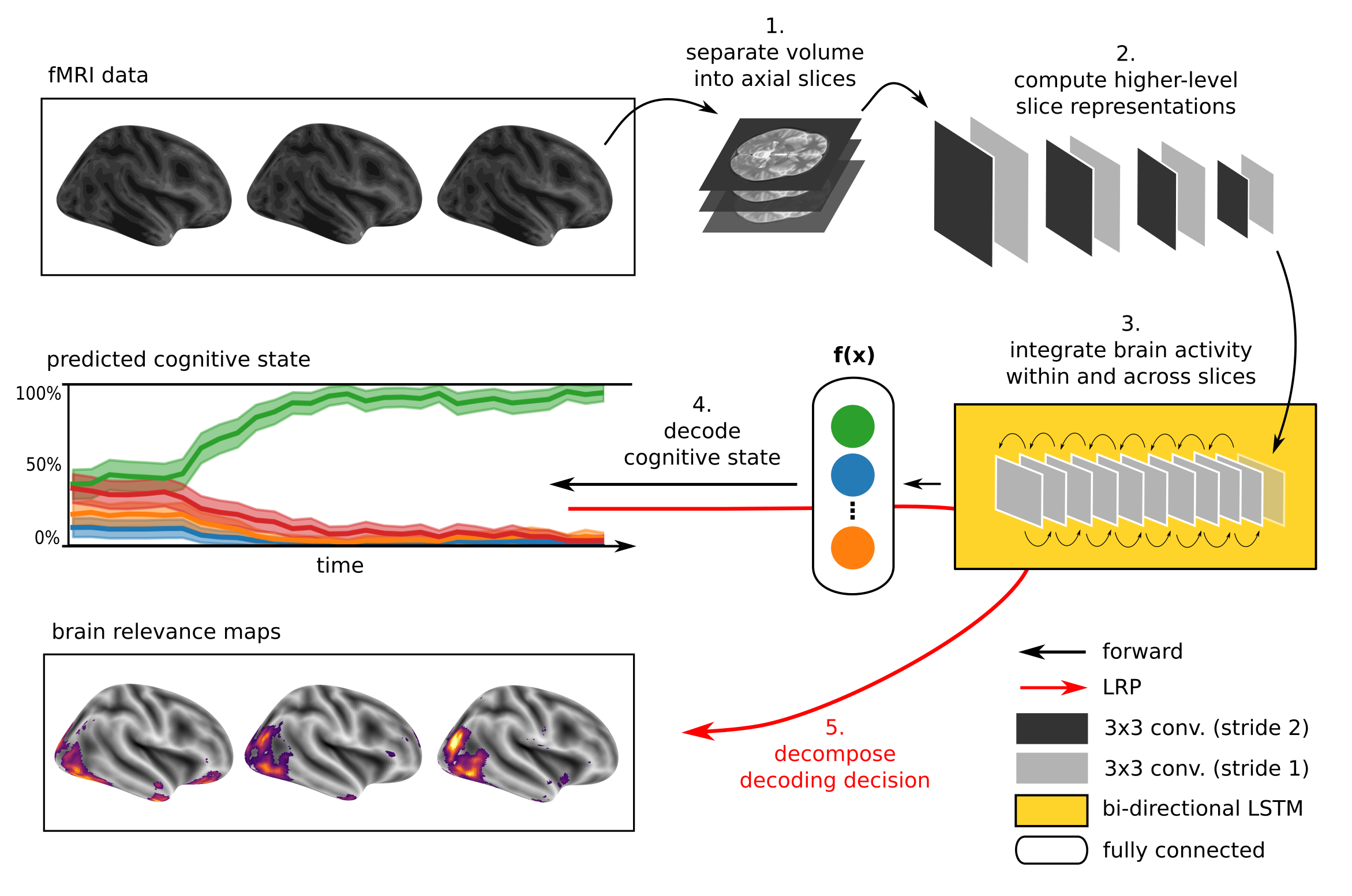}
 	\caption{Illustration of the DeepLight approach. Each whole-brain fMRI volume is sliced into a sequence of axial images. These images are then passed to a DL model consisting of a convolutional feature extractor, an LSTM and an output unit. First, the convolutional feature extractor reduces the dimensionality of the axial brain slices through a sequence of eight convolution layers. The resulting sequence of higher-level slice representations is then fed to a bi-directional LSTM, modeling the spatial dependencies of brain activity within and across brain slices. Lastly, the DL model outputs a decoding decision about the cognitive state underlying the fMRI volume, through a softmax output layer with one output neuron per cognitive state in the data. Once the prediction is made, DeepLight utilizes the LRP method to decompose the prediction into the contributions (or relevance) of the single input voxels to the prediction. Thereby, enabling an analysis of the association between fMRI data and cognitive state.}
 	\label{fig:DeepLight_illustration}
\end{figure*}

\subsubsection{Deep learning model}
The DL model underlying DeepLight consists of three distinct computational modules, namely a feature extractor, an LSTM, and an output unit (for an overview, see Fig.\ \ref{fig:DeepLight_illustration}). First, DeepLight separates each fMRI volume into a sequence of axial brain slices. These slices are then processed by a convolutional feature extractor \cite{lecun1995convolutional}, resulting in a sequence of higher-level, and lower-dimensional, slice representations. These higher-level slice representations are fed to an LSTM \cite{hochreiter1997long}, integrating the spatial dependencies of the observed brain activity within and across axial brain slices. Lastly, the output unit makes a decoding decision, by projecting the output of the LSTM into a lower-dimensional space, spanning the cognitive states in the data. Here, a probability for each cognitive state is estimated, indicating whether the input fMRI volume belongs to each of these states. This combination of convolutional and recurrent DL elements is inspired by previous research, showing that it is generally well-suited to learn the temporo-spatial dependency structure of long sequences of input data \cite{donahue2015long, mclaughlin2016recurrent, marban2018recurrent}. Importantly, the DeepLight approach is not dependent on any specific architecture of each of these three modules. The DL model architecture described in the following is exemplary and derived from previous work \cite{marban2018recurrent}. Further research is needed to explore the effect of specific module architectures on the performance of DeepLight.

The feature extractor used here was composed of a sequence of eight convolution layers \cite{lecun1995convolutional}. A convolution layer consists of a set of kernels (or filters) $w$ that each learn local features of the input image $a$. These local features are then convolved over the input, resulting in an activation map $h$, indicating whether a feature is present at each given location of the input: 

\begin{equation}
	h_{i,j} = g(\sum_{k=1}^{m} \sum_{l=1}^{m}(w_{k,l}a_{i+k+1, j+l-1}) + b)
	\label{eq:1} 
\end{equation}

Here, $b$ represents the bias of the kernel, while $g$ represents the activation function. $k$ and $l$ represent the row and column index of the kernel matrix, whereas $i$ and $j$ represent the row and column index of the activation map. 

Generally, lower-level convolution kernels (that are close to the input data) have small receptive fields and are only sensitive to local features of small patches of the input data (e.g., contrasts and orientations). Higher-level convolution kernels, on the other hand, act upon a higher-level representation of the input data, which has already been transformed by a sequence of preceding lower-level convolution kernels. Higher-level kernels thereby integrate the information provided by lower-level convolution kernels, allowing them to identify larger and more complex patterns in the data. 
We specified the sequence of convolution layers as follows (see Fig.\ \ref{fig:DeepLight_illustration}): conv3-16, conv3-16, conv3-16, conv3-16, conv3-32, conv3-32, conv3-32, conv3-32 (notation: conv(kernel size) - (number of kernels)). All convolution kernels were activated through a rectified linear unit function:

\begin{equation}
	g(z) = \max{(0, z)}
\end{equation}

Importantly, all kernels of the even-numbered convolution layers were moved over the input image with a stride size of one pixel and all kernels of odd-numbered layers with a stride size of two pixels. The stride size determines the dimensionality of the outputted slice representation. An increasing stride indicates more distance between the application of the convolution kernels to the input data. Thereby, reducing the dimensionality of the output representation at the cost of a decreasing sensitivity to differences in the activity patterns of neighbouring voxels. Yet, the activity patterns of neighbouring voxels are known to be highly correlated, leading to an overall low risk of information loss through a reasonable increase in stride size. This sequence of eight convolution layers resulted in a 960-dimensional representation of each volume slice.

To integrate the information provided by the resulting sequence of slice representations into a higher-level representation of the observed whole-brain activity, DeepLight applies a bi-directional LSTM \cite{hochreiter1997long}, containing two independent LSTM units. Each of the two LSTM units iterates through the entire sequence of input slices, but in reverse order (one from bottom-to-top and the other from top-to-bottom). An LSTM unit contains a hidden cell state $C$, storing information over an input sequence of length $S$ with elements $a_{s}$ and outputs a vector $h_{s}$ for each input at sequence step $s$. The unit has the ability to add and remove information from $C$ through a series of gates. In a first step, the LSTM unit decides what information from the cell state $C$ is removed. This is done by a fully-connected logistic layer, the forget gate $f$:

\begin{equation}
 f_{t} = \sigma(W_{f}[h_{s-1}, a_{s}] + b_{f})
 \label{eq:3} 
\end{equation}

Here, $\sigma$ indicates the logistic function (see eq. \ref{eq:logit}) and $[W, b]$ the gate's coefficients and bias. The forget gate outputs a number between 0 and 1 for each entry in the cell state $C$ at the previous sequence step $s-1$. Next, the LSTM unit decides what information is going to be stored in the cell state. This operation contains two elements: the input gate $i$, which decides which values of $C_{s}$ will be updated, and a $tanh$ layer, which creates a new vector of candidate values $C'_{s}$:

\begin{equation}
	i_{s} = \sigma(W_{i}[h_{s-1}, a_{s}] + b_{i})
	\label{eq:5} 
\end{equation}

\begin{equation}
	C'_{s} = \tanh(W_{c}[h_{s-1}, a_{s}] + b_{c})
	\label{eq:6} 
\end{equation}

\begin{equation}
	\tanh(z) = \frac{e^{z}-e^{-z}}{e^{z}+e^{-z}}
\end{equation}

Subsequently, the old cell state $C_{s-1}$ is updated into the new cell state $C_{s}$:

\begin{equation}
	C_{s} = f_{s} \cdot C_{s-1} + i_{s} \cdot C'_{s}
\label{eq:8} 
\end{equation}

Lastly, the LSTM computes its output $h_{s}$. Here, the output gate $o$, decides what part of $C_{s}$ will be outputted. Subsequently, $C_{s}$ is multiplied by another $tanh$ layer to make sure that $h_{s}$ is scaled between -1 and 1:

\begin{equation}
	o_{s} = \sigma(W_{o}[h_{s-1}, a_{s}] + b_{o})
	\label{eq:9} 
\end{equation}

\begin{equation}
	h_{s} = o_{s} \cdot \tanh(C_{s})
\label{eq:10}
\end{equation}

Each of the two LSTM units in our DL model contained 40 output neurons. To make a decoding decision, both LSTM units pass their output for the last sequence element to a fully-connected softmax output layer. The output unit contains one neuron per cognitive state in the data and assigns a probability to each of the $K$ (here, $K=4$) states, indicating the probability that the current fMRI sample belongs to this state:

\begin{equation}
	\sigma(z)_{j} = \frac{e^{z_{j}}}{\sum_{k=1}^{K}e^{z_{k}}} \mathrm{, \ with \ }j = 1, ..., K
	\label{eq:11} 
\end{equation}

\subsubsection{Layer-Wise Relevance Propagation in the DeepLight framework}
\label{lrp_in_deeplight}
To relate the decoded cognitive state and brain activity, DeepLight utilizes the Layer-Wise Relevance Propagation (LRP) method \cite{BachPLOS15, MonPR17, LapNCOMM19}. The goal of LRP is to identify the contribution of a single dimension $d$ of an input $a$ (with dimensionality $D$) to the prediction $f(a)$ that is made by a linear or non-linear classifier $f$. We denote the contribution of a single dimension as its relevance $R_{d}$. One way of decomposing the prediction $f(a)$ is by the sum of the relevance values of each dimension of the input:

\begin{equation}
	f(a) \approx \sum_{d=1}^{D} R_{d}
    \label{eq:12} 
\end{equation}

Qualitatively, any $R_{d} < 0$ can be interpreted as evidence against the presence of a classification target, while $R_{d} > 0$ denotes evidence for the presence of the target. Importantly, LRP assumes that $f(a) > 0$  indicates evidence for the presence of a target. 

Let's assume the relevance $R_{j}^{(l)}$ of a neuron $j$ at network layer $l$ for the prediction $f(a)$ is known. We would like to decompose this relevance into the messages $R_{i\leftarrow j}^{(l-1, l)}$ that are sent to those neurons $i$ in layer $l-1$ which provide the inputs to neuron $j$:

\begin{equation}
	R_{j}^{(l)} = \sum_{i\epsilon(l)} R_{i\leftarrow j}^{(l-1,l)}
\end{equation}

While the relevance of the output neuron at the last layer $L$ is defined as $R_{d}^{(L)} = f(a)$, the dimension-wise relevance scores on the input neurons are given by $R_{d}^{(1)}$.
For all weighted connections of the DL model in between (see eqs.\ \ref{eq:1}, \ref{eq:3}, \ref{eq:5}, \ref{eq:6} and \ref{eq:9}), DeepLight defines the messages $R_{i\leftarrow j}^{(l-1,l)}$ as follows:

\begin{equation}
	R_{i\leftarrow j}^{(l-1,l)} = \frac{z_{ij}}{z_{j} + \epsilon \cdot sign(z_{j})} R_{j}^{(l)}
    \label{eq:15}
\end{equation}

Here, $z_{ij} = a_{i}^{(l-1)}w_{ij}^{(l-1,l)}$ ($w$ indicating the coefficient weight and $a$ the input) and $z_{j}=\sum_{i}z_{ij}$, while $\epsilon$ represents a stabilizer term that is necessary to avoid numerical degenerations when $z_{j}$ is close to 0 (we set $\epsilon=0.001$).
 	
Importantly, the LSTM also applies another type of connection, which we refer to as multiplicative connection (see eqs.\ \ref{eq:8} and \ref{eq:10}). Let $z_{j}$ be an upper-layer neuron whose value in the forward pass is computed by multiplying two lower-layer neuron values $z_{g}$ and $z_{s}$ such that $z_{j}=z_{g} \cdot z_{s}$. These multiplicative connections occur when we multiply the outputs of a $gate$ neuron, whose values range between 0 and 1, with an instance of the hidden cell state, which we will call $source$ neuron. For these types of connections, we set the relevances of the gate neuron $R_{g}^{(l-1)}=0$ and the relevances of the source neuron $R_{s}^{(l-1)}=R_{j}^{(l)}$, where $R_{j}^{(l)}$ denotes the relevances of the upper layer neuron $z_{j}$ (as proposed in \cite{ArrWASSA17}). The reasoning behind this rule is that the gate neuron already decides in the forward pass how much of the information contained in the source neuron should be retained to make the classification. Even if this seems to ignore the values of the neurons $z_{g}$ and $z_{s}$ for the redistribution of relevance, these are actually taken into account when computing the value $R_{j}^{(l)}$ from the relevances of the next upper-layer neurons to which $z_{j}$ is connected by the weighted connections. We refer the reader to \cite{SamITU18b, MonDSP18} for more information about explanation methods. 

In the context of this work, we decomposed the predictions of DeepLight for the actual cognitive state underlying each fMRI sample, as we were solely interested in understanding what DeepLight used as evidence in favor of the presence of this state. We also restricted the LRP analysis to those brain samples that the DL model classified correctly, because we can only assume that the DL model has learned a meaningful mapping between brain data and cognitive state, if it is able to accurately decode the cognitive state.

\subsubsection{DeepLight training}
We iteratively trained DeepLight through backpropagation \cite{rumelhart1986learning} over a period of 60 epochs by the use of the ADAM optimization algorithm as implemented in tensorflow 1.4 \cite{abadi2016tensorflow}. To prevent overfitting, we applied dropout regularization to all network layers \cite{srivastava2014dropout}, global gradient norm clipping (with a clipping threshold of 5) \cite{pascanu2013difficulty}, as well as an early stopping of the training (see Supplementary Fig.\ S2 for an overview of training statistics). During the training, we set the dropout probability to 50\% for all network layers, except for the first four convolution layers, where we reduced the dropout probability to 30\% for the first two layers and 40\% for the third and fourth layer. Each training epoch was defined as a complete iteration over all samples in the training dataset (see Section \ref{sec:fmri_preprocessing_details}). We used a learning rate of 0.0001 and a batch size of 32. All network weights were initialized by the use of a normal-distributed random initialization scheme \cite{glorot2010understanding}. The DL model was written in tensorflow 1.4 \cite{abadi2016tensorflow} and the interprettensor library (\href{https://github.com/VigneshSrinivasan10/interprettensor}{https://github.com/VigneshSrinivasan10/interprettensor}).

\subsection{Estimating brain maps}
\label{sec:classifier_training_details}
In the following, we will briefly describe the procedures that we used to estimate a set of brain maps for each cognitive state with each analysis approach. As our subsequently presented analyses are sub-divided into a separate analysis on the subject- and group-level (containing the data of an individual subject or the data of the entire group of subjects), we will also divide the following section according to the subject- and group-level. The data of many subjects (with approximately 1GB per subject) can easily exceed the working memory capacities of a regular working station. For this reason, we adapted the parameter estimation procedures for the searchlight analysis and whole-brain lasso, when switching from the subject- to the group-level.

\subsubsection{Subject-level}
\label{sec:classifier_training_details_subject}
Our GLM analyses included one predictor for each of the four cognitive states in the design matrix (each representing a box-car function for the occurrence of a cognitive state; for methodological details on the GLM, see Section \ref{sec:GLM_details}). We convolved these predictors with a canonical hemodynamic response function (HRF) \cite{glover1999deconvolution}, as implemented in NiPy 0.4.1 \cite{gorgolewski2011nipype}, to generate the model predictors. We added temporal derivative terms derived from each predictor, an intercept and an indicator of the experiment run to the design matrix, which we all treated as confounds of no interest. The derivative terms were computed by the use of the cosine drift model as implemented in NiPy 0.4.1. All $\beta$-coefficients and error terms of the GLM analysis were estimated by the use of a first-level autoregressive model, as implemented in NiPy 0.4.1. To generate a set of subject-level brain maps, we computed a linear first-level contrast within the data of each individual subject (representing a linear contrast between one of the cognitive states and all others). The resulting brain maps indicate the estimated Z-values of these contrasts.

To obtain a set of subject-level brain maps for each cognitive state with the searchlight analysis, we trained searchlight in a one-vs-rest procedure. Here, one SVM classifier is trained at each location in the brain to distinguish each cognitive state from all others. A decoding decision is then made according to the classifier with the most certainty that the sample belongs to its respective cognitive state. We first trained the searchlight analysis within the data of the first experiment run of a subject (see Section \ref{sec:experiment_paradigm_details}) and subsequently predicted the cognitive states underlying the data of the second experiment run. The resulting brain maps indicate the decoding accuracies achieved by each of these SVM classifiers in the second experiment run at each searchlight location in the brain. 

Similarly, we also trained the whole-brain lasso in a one-vs-rest procedure. To determine the magnitude of the regularization parameter $\lambda$, we additionally applied a grid search. First, we split the full training data of a subject, containing the data of the first experiment run (see Section \ref{sec:experiment_paradigm_details}), into the eight experiment blocks of this run (two per cognitive state). We then separated these blocks into a new training dataset (containing the first experiment block of each cognitive state) and a new validation dataset (containing the second experiment block of each cognitive state). Subsequently, for each value of the parameter grid, we fit the whole-brain lasso to the data of the newly formed training dataset and evaluated its performance on the new validation data. Importantly, we utilized a logistic model implementation of the scikit-learn python library \cite{abraham2014machine}. Here, the regularization parameter ($C$) is implemented inversely to the regularization strength $\lambda$ (with lower values indicating stronger regularization; see eq. \ref{eq:4}). With this procedure, we evaluated a grid of 100 logarithmically spaced $C$-values between 1e-6 and 100. From these values we then selected the $C$-parameter for the subject that achieved the highest decoding accuracy in the new validation dataset (for an overview of the selected subject $C$-parameters, see Supplementary Table 1). Subsequently, we used the selected $C$-value to fit the whole-brain lasso to the full training data of the subject (containing the entire data of the first experiment run). The resulting brain maps represent the coefficient estimates of each of these one-vs-rest logistic models.

Lastly, the subject-level brain maps of DeepLight were generated by decomposing the decoding decisions of DeepLight for each correctly classified fMRI sample of a subject with the LRP method (see Section \ref{sec:DeepLight_details}). Importantly, we restricted the LRP analysis to those fMRI samples that were collected 5 - 15s after the onset of the experiment block, as we expect the HRF to be strongest within this time period. To then aggregate the resulting set of relevance maps for each decomposed fMRI sample within each cognitive state, we smoothed each relevance map with a 3mm FWHM Gaussian kernel and averaged all relevance volumes belonging to a cognitive state, resulting in one brain map per subject and cognitive state.

\subsubsection{Group-level}
\label{sec:classifier_training_details_group}
To generate a set of group-level brain maps with the GLM, we computed a second-level GLM contrast by the use of the standard two-stage procedure for a random-effects group-level analysis, as proposed by Holmes \& Friston \cite{holmes1998generalisability}. Here, the subject-level regression coefficients $\beta$ (see Section \ref{sec:GLM_details}) are treated as random effects in a second-level linear contrast analysis, where the distribution of first-level $\beta$-contrasts is assessed by the use of a one-sample t-test (again, contrasts were computed between each cognitive state and all others). The resulting group-level brain maps represent the t-values of this test.

For the group-level searchlight analysis, we trained and evaluated the searchlight on the $\beta$-coefficient maps of a first-level GLM analysis of each individual subject (resulting in one $\beta$-coefficient map per subject and cognitive state; see Section \ref{sec:GLM_details}). This is a common approach for group-level predictions with the searchlight analysis and is widely applied in the neuroscience literature (e.g., \cite{weygandt2012diagnosing, helfinstein2014predicting, schuck2016human, reverberi2018neural}). First, we trained the searchlight analysis in a one-vs-rest procedure on the subject-level $\beta$-coefficient maps of the training dataset. Subsequently, we used each of the trained searchlight classifiers to decode the cognitive states underlying each subject-level $\beta$-coefficient map in the test data. The resulting group-level brain maps represent the decoding accuracies achieved by each of these searchlight classifiers in the test data.

On the group-level, we trained the whole-brain lasso in a stochastic gradient descent learning procedure \cite{kiefer1952stochastic}. Here, the regularized logistic model (see eq.\ \ref{eq:4}) is fit iteratively to subsets of the full training data. At each iteration, the gradient of the loss function is estimated and the model's parameters are updated accordingly. To determine the strength of the regularization parameter $\lambda$, we again applied a grid search procedure. For each value of the $\lambda$-grid, we trained the whole-brain lasso over 25 epochs. In each epoch, we randomly selected the fMRI data of five subjects from the training dataset. We then randomly drew 50 batches, each containing 50 randomly drawn TRs, and updated the whole-brain lasso parameters iteratively for each batch. After completing the 25 epochs, we evaluated the decoding performance of the whole-brain lasso on the full test dataset. Overall, we evaluated 20 different $\lambda$-parameters in this grid-search and selected the $\lambda$-value achieving the highest decoding accuracy in the test dataset ($\lambda$=0.0001; for an overview of the evaluated $\lambda$-values and resulting decoding accuracies, see Supplementary Table 2). We then used the selected $\lambda$-parameter to train the whole-brain lasso in the same stochastic gradient procedure described before. This time, however, spanning 200 training epochs. The group-level brain maps of the whole-brain lasso represent the resulting one-vs-rest logistic model coefficients.

Lastly, to generate a set of group-level brain maps with DeepLight, we averaged the subject-level brain maps for all subjects in the held-out test dataset within each cognitive state, resulting in one group-level brain map per cognitive state.

\section{Results}
\label{sec:results}

\subsection{DeepLight accurately decodes cognitive states from fMRI data}
\label{sec:results_decoding_comparison}

\begin{figure*}[!th]
  \centering
  \includegraphics[width=0.75\linewidth]{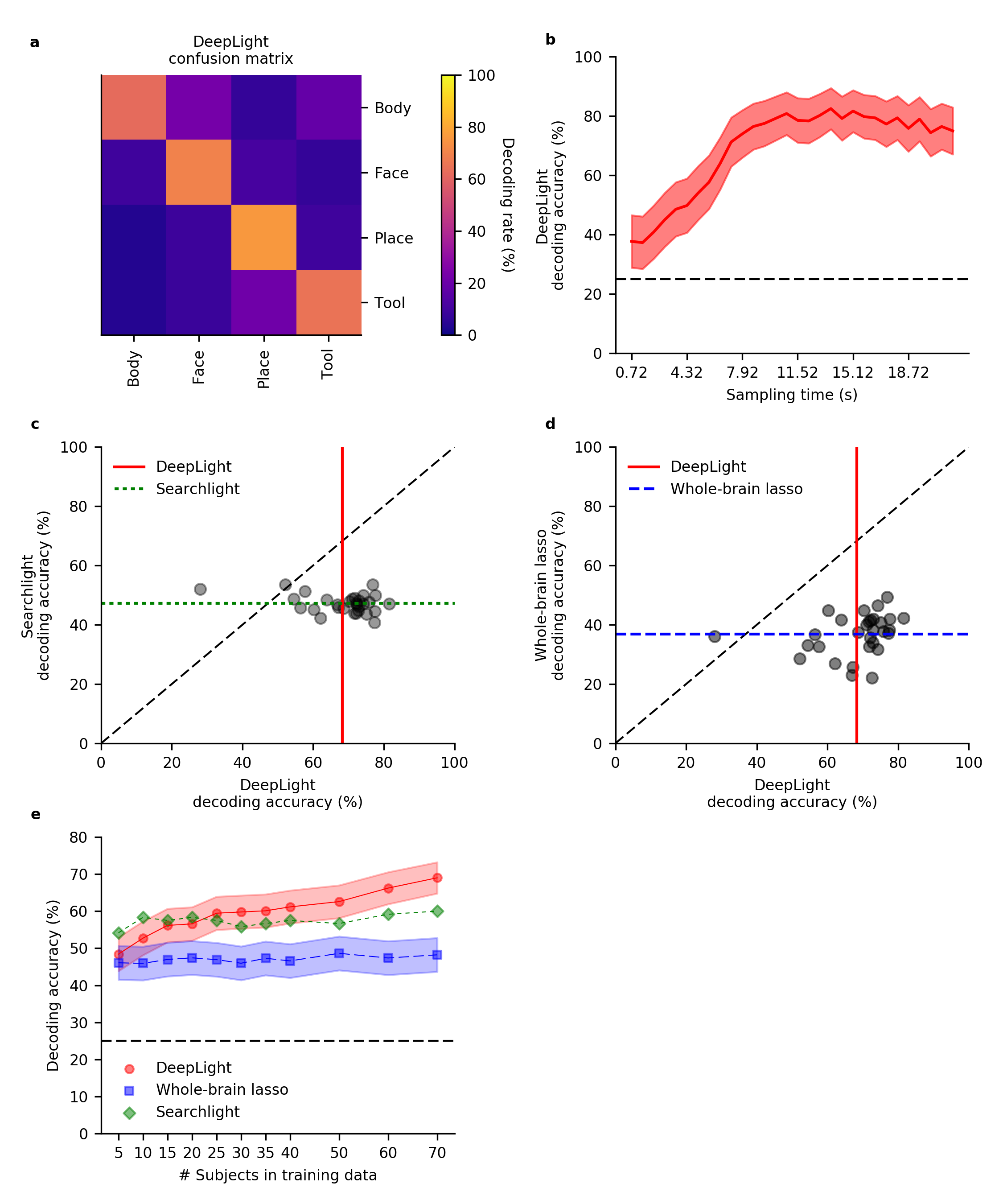}
  \caption{Evaluating DeepLight's performance in decoding the cognitive states from the held-out fMRI test data. \textbf{a}: Confusion matrix of DeepLight's decoding decisions. \textbf{b}: Average decoding performance of DeepLight over the course of an experiment block. \textbf{c-d}: Comparison of DeepLight's decoding performance (red) to the searchlight analysis (c; green) and whole-brain lasso (d; blue) on the level of individual subjects. The average decoding accuracy for a subject is indicated by a black scatter point. Colored lines indicate the average decoding accuracy across subjects. \textbf{e}: Average decoding accuracy of the searchlight (green), whole-brain lasso (blue) and DeepLight (red), when these are trained on a subset of the subjects contained in the original full training dataset. Black dashed horizontal lines indicate chance level.} 
  \label{fig:decoding_performance}
\end{figure*}

A key prerequisite for the DeepLight analysis (as well as all other decoding analyses) is that it achieves reasonable performance in the decoding task at hand. Only then we can assume that it has learned a meaningful mapping from the fMRI data to the cognitive states and interpret the resulting brain maps as informative about these states.

Overall, DeepLight accurately decoded the cognitive states underlying 69\% of the fMRI samples in the held-out test dataset (61.1\%, 73.6\%, 74.5\%, 66.7\% for body, face, place and tool respectively; Fig.\ \ref{fig:decoding_performance}\ a). It generally performed best at discriminating the body and place (5.1\% confusion in the held-out data), face and tool (7.8\% confusion in the held-out data), body and tool (9.5\% confusion in the held-out data) and face and place (10.4\% confusion in the held-out data) stimuli from the fMRI data, while it did not perform as well in discriminating place and tool and body and face stimuli (15\% confusion in the held-out data respectively).

Note that DeepLight's performance in decoding the four cognitive states from the fMRI data varied over the course of an experiment block (Fig.\ \ref{fig:decoding_performance}\ b). DeepLight performed best in the middle and later stages of the experiment block, where the average decoding accuracy reaches 80\%. This finding is generally in line with the temporal evolution of the hemodynamic response function (HRF; \cite{lindquist2009modeling}) measured by the fMRI (the HRF is known to be strongest 5-10 seconds after to the onset of the underlying neuronal activity). 

To further evaluate DeepLight's performance in decoding the cognitive states from the fMRI data, we compared its performance in decoding these states, on the subject- and group-level, to the searchlight analysis and whole-brain lasso. 

\subsubsection{Subject-level}
For the subject-level comparison, we first trained both, the searchlight analysis and whole-brain lasso on the fMRI data of the first experiment run of a subject from the held-out test dataset (for an overview of the estimation procedures, see Section \ref{sec:classifier_training_details_subject}). We then used the data of the second experiment run of the same subject to evaluate their decoding performance (by predicting the cognitive states underlying each fMRI sample of the second experiment run). Importantly, we also decoded the same fMRI samples with DeepLight. Note that DeepLight, in comparison to the other approaches, did not see any data of the subject during the training, as it was solely trained on the data of the 70 subjects in the training dataset (see Section \ref{sec:experiment_paradigm_details}).

DeepLight clearly outperformed the other decoding approaches, by decoding the cognitive states more accurately for 28 out of 30 subjects, when compared to the searchlight analysis (while the searchlight analysis achieved an average decoding accuracy of 47.2\% across subjects, DeepLight improved upon this performance by 22.4\%, with an average decoding accuracy of 69.3\%, t(29)= 11.28, p<0.0001; Fig.\ \ref{fig:decoding_performance}\ c), and for 29 out of 30 subjects, when compared to the whole-brain lasso (while the whole-brain lasso achieved an average decoding accuracy of 37\% across subjects, DeepLight improved upon this performance by 32\%; t(29)=15.74, p<0.0001; Fig.\ \ref{fig:decoding_performance}\ d). To further ascertain that the observed differences in decoding performance between the searchlight and DeepLight did not result from the linearity contained in the Support Vector Machine (SVM; \cite{cortes1995support}) of the the searchlight analysis, we replicated our subject-level searchlight analysis, by the use of a non-linear radial basis function kernel (RBF; \cite{cortes1995support, muller2001introduction, scholkopf2002learning}) SVM (Supplementary Fig. S3). However, the decoding accuracies achieved by the RBF-kernel SVM were not meaningfully different from those of the linear-kernel SVM (t(29)=-1.75, p=0.09).

\subsubsection{Group-level}
For the group-level comparison, we trained the searchlight analysis and whole-brain lasso on the data of all 70 subjects contained in the training dataset (for details on the estimation procedures, see Section \ref{sec:classifier_training_details_group}). Subsequently, we evaluated their performance in decoding the cognitive states in the full held-out test data. 

Again, DeepLight clearly outperformed the other approaches in decoding the cognitive states. While the searchlight analysis achieved an average decoding accuracy of 60\% and the whole-brain lasso an average decoding accuracy of 48\%, DeepLight improved upon these performances by 9\% (t(29)=5.80, p<0.0001) and 21\% (t(29)=13.39, p<0.0001) respectively.

A key premise of DL methods, when compared to more traditional decoding approaches, is that their decoding performance improves better with growing datasets. To test this, we repeatedly trained all three decoding approaches on a subset of the training dataset (including the data of 5, 10, 15, 20, 25, 30, 35, 40, 50, 60 and 70 subjects), and validated their performance at each iteration on the full held-out test data (Fig.\ \ref{fig:decoding_performance}\ e). Overall, the decoding performance of DeepLight increased by 0.27\% (t(10)=10.9, p<0.0001) per additional subject in the training dataset, whereas the performance of the whole-brain lasso increased by 0.03\% (t(10)=3.02, p=0.015) and the performance of the searchlight analysis only marginally increased by 0.04\% (t(10)=2.08, p=0.067). Nevertheless, the searchlight analysis outperformed DeepLight in decoding the cognitive states from the data when only little training data were available (here, 10 or less subjects (t(29)=-4.39, p<0.0001). The decoding advantage of DeepLight, on the other hand, came to light when the data of 50 or more subjects were available in the training dataset (t(29)=3.82, p=0.0006). DeepLight consistently outperformed the whole-brain lasso, when it was trained on the data of at least 10 subjects (t(29)=5.32, p=0.0045).

\subsection{DeepLight identifies physiologically appropriate associations between cognitive states and brain activity}
\label{sec:results_group_level_maps}

\begin{sidewaysfigure*}[p]
    \vspace{-1pt}
    \hspace{-1pt}
    \makebox[\linewidth]{
        \includegraphics[width=0.9\linewidth]{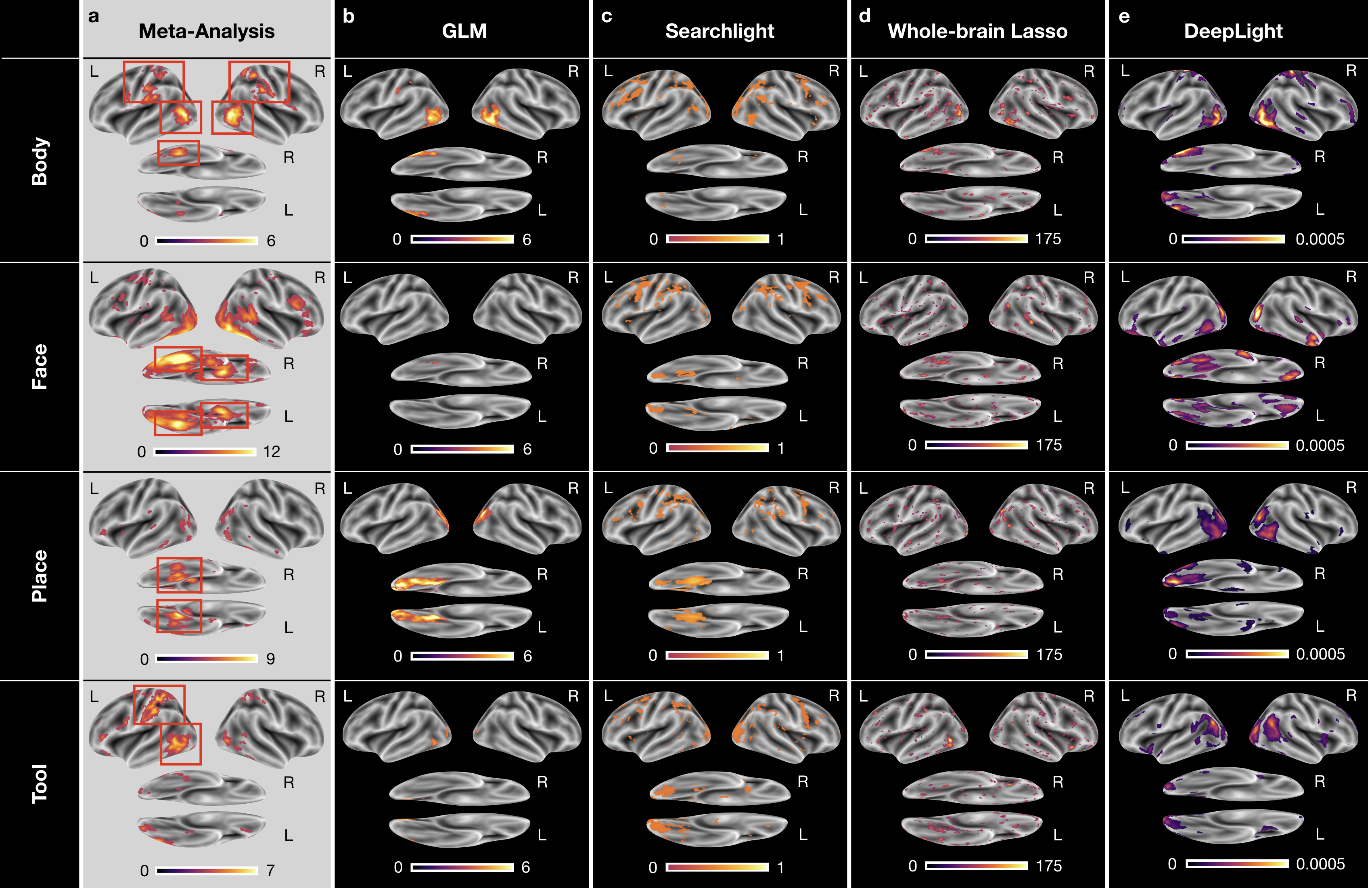}}
    \caption{Group-level brain maps for each cognitive state and analysis approach: \textbf{a}: Results of a NeuroSynth meta-analysis for the terms "body", "face", "place" and "tools". The brain maps were thresholded at a false-discovery rate of 0.01. Red boxes point out the regions-of-interest for each cognitive state. \textbf{b}: Results of the GLM group-level analysis. The brain maps were thresholded at a false-discovery rate of 0.1. \textbf{c-e}: Results of the group-level searchlight analysis (c), whole-brain lasso (d) and DeepLight (e). The brain maps were each thresholded at the 90th percentile of their values. All brain maps were projected onto the FsAverage5 surface template \cite{fischl2012freesurfer}.}
    \label{fig:group_maps}
\end{sidewaysfigure*}

Our previous analyses have shown that DeepLight has learned a meaningful mapping between the fMRI data and cognitive states, by accurately decoding these states from the data. Next, we therefore tested DeepLight's ability to identify the brain areas associated with the cognitive states, by decomposing its decoding decisions with the LRP method (see Section \ref{sec:classifier_training_details}). Subsequently, we compared the resulting brain maps of DeepLight to those of the GLM, searchlight analysis and whole-brain lasso. Again, we sub-divided this comparison into a separate analysis on the subject- and group-level. 

To evaluate the quality of the brain maps resulting from each analysis approach, we performed a meta-analysis of the four cognitive states with NeuroSynth (for details on NeuroSynth, see Supplementary Information Section 1 and ref. \cite{yarkoni2011large}). NeuroSynth provides a database of mappings between cognitive states and brain activity, based on the empirical neuroscience literature. Particularly, the resulting brain maps used here indicate whether the probability that an article reports a specific brain activation is different, when it includes a specific term (e.g., "face") compared to when it does not. With this meta-analysis, we defined a set of \textit{regions-of-interest} (ROIs) for each cognitive state (as defined by the terms "body", "face", "place", and "tools"), in which we would expect the various analysis approaches to identify a positive association between the cognitive state and brain activity (for an overview, see Fig.\ \ref{fig:group_maps}\ a). These ROIs were defined as follows: the upper parts of the middle and inferior temporal gyrus, the postcentral gyrus, as well as the right fusiform gyrus for the body state, the fusiform gyrus (also known as the fusiform face area FFA; \cite{haxby2001distributed, heekeren2004general}) and amygdala for the face state, the parahippocampal gyrus (or parahippocampal place area PPA; \cite{haxby2001distributed, heekeren2004general}) for the place state and the upper left middle and inferior temporal gyrus as well as the left postcentral gyrus for the tool state. To ensure comparability with the results of the meta-analysis, we restricted all analyses of brain maps to the estimated positive associations between brain activity and cognitive states (e.g., positive relevance values as well as positive GLM and whole-brain lasso coefficients). A negative Z-value in the meta-analysis indicates a lower probability that an article reports a specific brain activation when it includes a specific term, compared to when it does not include the term. A negative value in the meta-analysis is therefore conceptually different to negative values in the brain maps of our analyses (e.g., negative relevance values or negative whole-brain lasso coefficients). These can be interpreted as evidence against the presence of a cognitive state or as evidence for the presence of any of the other cognitive states in our dataset.

\subsubsection{Group-level}
\label{sec:results_group_level_maps_group}
To determine the voxels that each analysis approach associated with a cognitive state, we defined a threshold for the values of each group-level brain map, indicating those voxels that are associated most strongly with the cognitive state. For the GLM analysis, we thresholded all P-values at a false-discovery-rate of 0.1 (Fig.\ \ref{fig:group_maps}\ b). Similarly, for all decoding analyses, we thresholded each brain map at the 90th percentile of its values (Fig.\ \ref{fig:group_maps}\ c-e). For the whole-brain lasso and DeepLight, the remaining 10 percent of values indicate those brain regions whose activity these approaches generally weight most in their decoding decisions. For the searchlight analysis, the remaining 10 percent of values indicate those brain regions in which the searchlight analysis achieved the highest decoding accuracy. Due to the strong sparsity of the group-level brain maps of the whole-brain lasso, we additionally smoothed the respective maps with a 3mm FWHM Gaussian kernel. 

All analysis approaches correctly associated activity in the upper parts of the middle and inferior temporal gyrus with body stimuli. The GLM, whole-brain lasso and DeepLight also correctly associated activity in the right fusiform gyrus with body stimuli. Only DeepLight correctly associated activity in the postcentral gyrus with these stimuli. The GLM, whole-brain lasso and DeepLight further all correctly associated activity in the right FFA with face stimuli. None of the approaches, however, associated activity in the left FFA with face stimuli. Interestingly, the searchlight analysis did not associate the FFA with face stimuli at all. All analysis approaches also correctly associated activity in the PPA with place stimuli. Lastly, for tool stimuli, the GLM and whole-brain lasso correctly associated activity in the left inferior temporal sulcus with stimuli of this class. The searchlight analysis and whole-brain lasso only did so marginally. None of the approaches associated activity in the left postcentral gyrus with tool stimuli. 

Overall, DeepLight's group-level brain maps accurately associated each of the ROIs with their respective cognitive states. Interestingly, DeepLight also associated a set of additional brain regions with the face and tool stimulus classes that were not identified by the other analysis approaches (see Fig.\ \ref{fig:group_maps}\ e). For face stimuli, these regions are the orbitofrontal cortex and temporal pole. While the temporal pole has been shown to be involved in the ability of an individual to infer the desires, intentions and beliefs of others (\textit{theory-of-mind}; for a detailed review, see \cite{olson2007enigmatic}), the orbitofrontal cortex has been associated with the processing of emotions in the faces of others (for a detailed review, see \cite{adolphs2002neural}). For tool stimuli, DeepLight additionally utilized the activity of the temporoparietal junction (TPJ) to decode these stimuli. The TPJ has been shown to be associated with the ability of an individual to discriminate self-produced actions and the actions produced by others and is generally regarded of as a central hub for the integration of body-related information (for a detailed review, see \cite{decety2006power}). Although it is not clear why only DeepLight associated these brain regions with the face and tool stimulus classes, their assumed functional roles do not contradict this association.

\subsubsection{Subject-level}
\label{sec:results_group_level_maps_subject}
The goal of the subject-level analysis was to test the ability of each analysis approach to identify the physiologically appropriate associations between brain activity and cognitive state on the level of each individual.

To quantify the similarity between the subject-level brain maps and the results of the meta-analysis, we defined a similarity measure. Given a target brain map (e.g., the results of our meta-analysis), this measure tests for each voxel in the brain whether a source brain map (e.g., the results of our subject-level analyses) correctly associates this voxel's activity with the cognitive state (true positive), falsely associates the voxel's activity with the cognitive state (false positives) or falsely does not associate the voxel's activity with the cognitive state (false negatives). Particularly, we derived this measure from the well-known F1-score in machine learning (see Supplementary Information Section 2 and ref. \cite{goutte2005probabilistic}). The benefit of the F1-score, when compared to simply computing the ratio of correctly classified voxels in the brain, is that it specifically considers the brain map's precision and recall and is thereby robust to the overall size of the ROIs in the target brain map. Here, precision describes the fraction of true positives from the total number of voxels that are associated with a cognitive state in the source brain map. Recall, on the other hand, describes the fraction of true positives from the overall number of voxels that are associated with a cognitive state in the target brain map. Generally, an F1-score of 1 indicates that the brain map has both, perfect precision and recall with respect to the target, whereas the F1-score is worst at 0. 

To obtain an F1-score for each subject-level brain map, we again thresholded each individual brain map. For the GLM, we defined all voxels with a P-value greater than 0.005 (uncorrected) as not associated with the cognitive state and all others as associated with the cognitive state. For the searchlight analysis, whole-brain lasso and DeepLight, we defined all voxels with a value below the 90th percentile of the values within the brain map as not associated with the cognitive state and all others as associated with the cognitive state. To additionally compensate for the smoothing that was applied to the subject-level relevance maps of the DeepLight analysis (see Section \ref{sec:classifier_training_details_subject}), we also smoothed the subject-level brain maps of the GLM, searchlight analysis and whole-brain lasso with a 3mm FWHM Gaussian kernel.

Overall, DeepLight's subject-level brain maps had meaningfully larger F1-scores for the body and face stimulus classes, when compared to those of the GLM (t(29)=9.50, p<0.0001 for body stimuli, Supplementary Fig.\ S4\ a; t(29)=12.58, p<0.0001 for face stimuli, Supplementary Fig.\ S4\ d), searchlight analysis (t(29)=10.30, p<0.0001 for body stimuli, Supplementary Fig.\ S4\ b; t(29)=4.97, p<0.0001 for face stimuli, Supplementary Fig.\ S4\ e), and whole-brain lasso (t(29)=3.40, p=0.002 for body stimuli, Supplementary Fig.\ S4\ c; t(29)=17.03, p<0.0001 for face stimuli, Supplementary Fig.\ S4\ f). Similarly, for place stimuli, the F1-scores of DeepLight's subject-level brain maps were meaningfully larger, when compared to the GLM (t(29)=7.86, p<0.0001, Supplementary Fig.\ S4\ g) and whole-brain lasso (t(29)=4.32, p=0.0002, Supplementary Fig.\ S4\ i), while not being meaningfully different from those of the searchlight analysis (t(29)==1.10, p=0.28 Supplementary Fig.\ S4\ h). For tool stimuli, the GLM, searchlight and whole-brain lasso generally achieved higher subject-level F1-scores than DeepLight (t(29)=-7.70, p<0.0001, Supplementary Fig.\ S4\ j; t(29)=-4.22, p=0.0002, Supplementary Fig.\ S4\ k; t(29)=2.23, p=0.034, Supplementary Fig.\ S4\ l for the GLM, searchilght and whole-brain lasso respectively). This was to expect, however, due to the DeepLight's focus on the TPJ, when decoding stimuli of this type.

To ascertain that the results of this comparison were not dependent on the thresholds that we chose, we replicated the comparison for each combination of the 85th, 90th and 95th percentile threshold for the brain maps of the searchlight analysis, whole-brain lasso and DeepLight, as well as a P-threshold of 0.05, 0.005, 0.0005 and 0.00005 for the brain maps of the GLM. Within all combinations of percentile values and P-thresholds, the presented results of the F1-comparison were generally stable (see Supplementary Table 3-6).

\subsection{DeepLight accurately identifies physiologically appropriate associations between cognitive states and brain activity on multiple levels of data granularity}
\label{sec:results_granularity_analysis}

\begin{figure*}[!h]
  \centering
  \includegraphics[width=\linewidth]{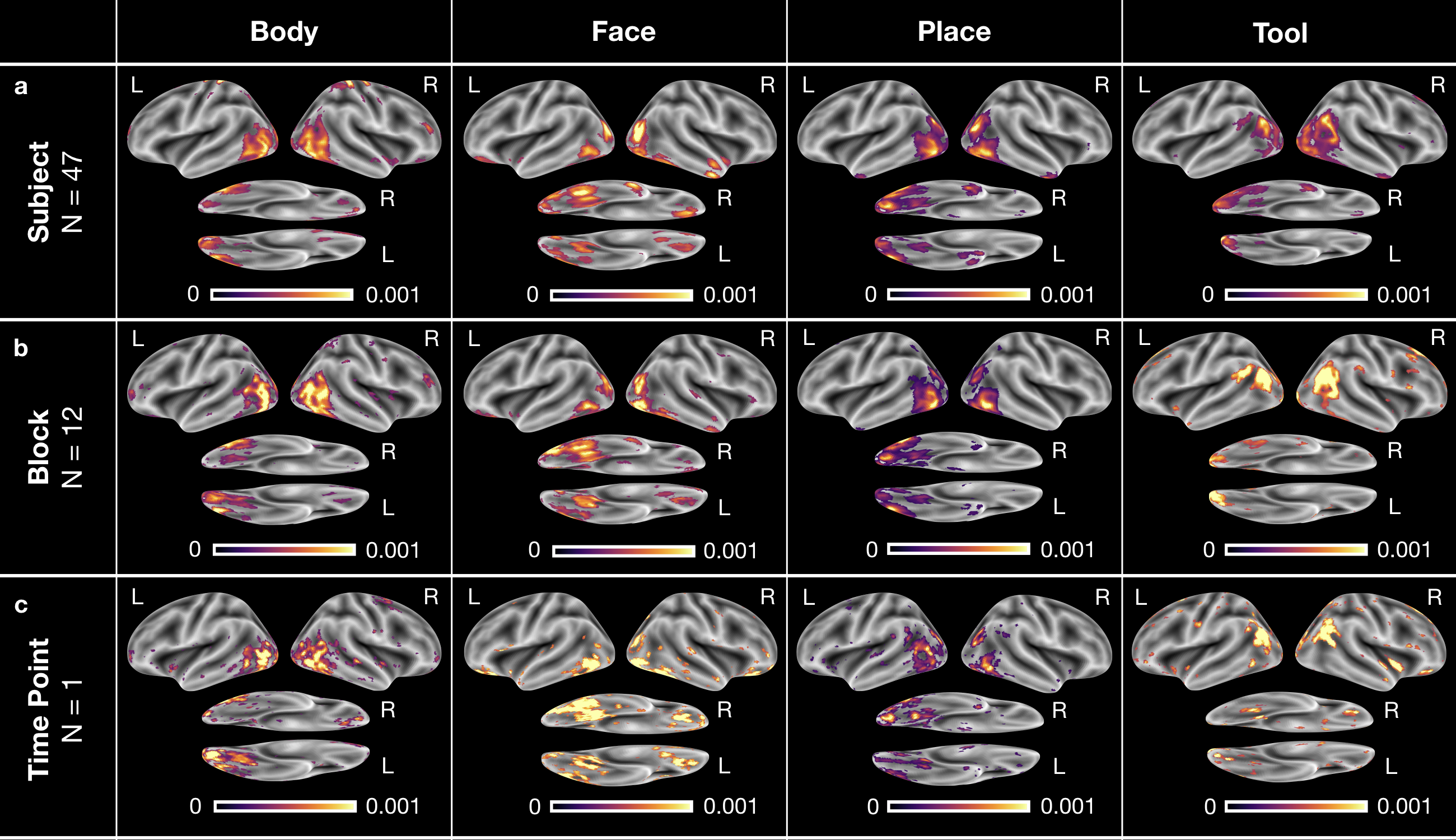}
  \caption{Exemplary DeepLight brain maps for each of the four cognitive states on different levels of data granularity for a single subject. All brain maps belong to the subject with the highest decoding accuracy in the held-out test dataset. \textbf{a}: Average relevance maps for all correctly classified TRs of the subject (with an average of 47 TRs per cognitive state). \textbf{b}: Average relevance maps for all correctly classified TRs of the first experiment block of each cognitive state in the first experiment run (with an average of 12 TRs per cognitive state). \textbf{c}: Exemplar relevance maps for a single TR of the first experiment block of each cognitive state in the first experiment run. All relevance maps were thresholded at the 90th percentile of their values and projected onto the FsAverage5 surface template \cite{fischl2012freesurfer}.}
  \label{fig:granularity_analysis}
\end{figure*}

DeepLight's ability to correctly identify the physiological appropriate associations between cognitive states and brain activity is exemplified in Figure \ref{fig:granularity_analysis}. Here, the distribution of relevance values for the four cognitive states is visualized on three different levels of data granularity of an exemplar subject (namely, the subject with the highest decoding accuracy in Fig.\ \ref{fig:decoding_performance}\ c-d): First, on the level of the overall distribution of relevance values of each cognitive state of this subject (Fig.\ \ref{fig:granularity_analysis}\ a; incorporating an average of 47 TRs per cognitive state), then on the level of the first experiment block of each cognitive state in the first experiment run (Fig.\ \ref{fig:granularity_analysis}\ b; incorporating an average of 12 TRs per cognitive state) and lastly on the level of a single brain sample of each cognitive state (Fig.\ \ref{fig:granularity_analysis}\ c; incorporating a single TR per cognitive state).

On all three levels, DeepLight utilized the activity of a similar set of brain regions to identify each of the four cognitive states. Importantly, these regions largely overlap with those identified by the DeepLight group-level analysis (Fig.\ \ref{fig:group_maps}\ e) as well as the results of the meta-analysis (Fig.\ \ref{fig:group_maps}\ a).

\subsection{DeepLight's relevance patterns resemble temporo-spatial variability of brain activity over sequences of single fMRI samples}
\label{sec:results_ffa_ppa_block_anlaysis}

\begin{figure*}[!t]
  \centering
  \includegraphics[width=\linewidth]{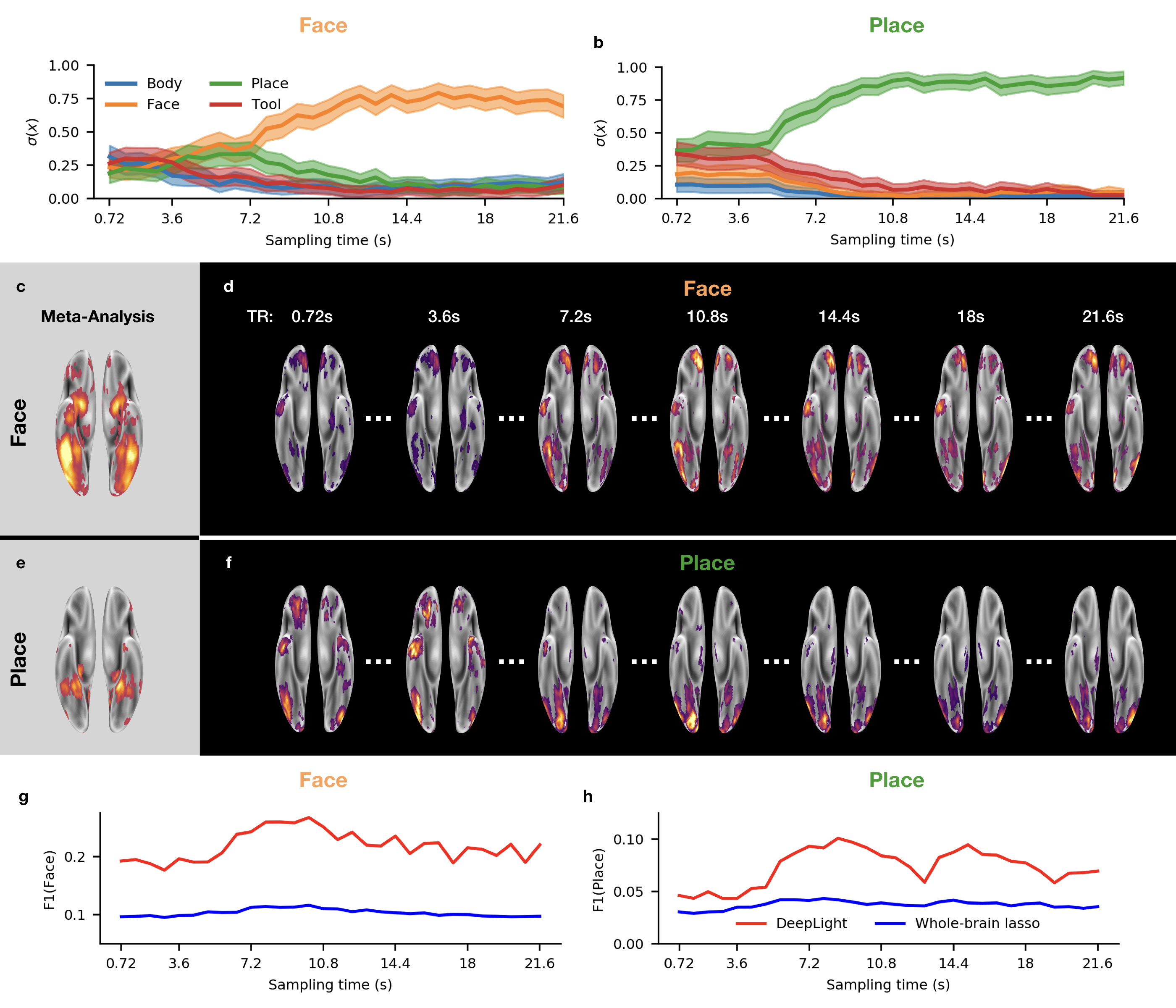}
  \caption{DeepLight analysis of the temporo-spatial distribution of brain activity in the first experiment block of the face and place stimulus classes in the second experiment run of the held-out test dataset. \textbf{a-b}: Average predicted probability that each fMRI volume belongs to each of the four cognitive states. \textbf{c \& e}: Results of a meta-analysis with the NeuroSynth database for the face and place stimulus classes (for details on the meta-analysis, see Supplementary Information Section 1). \textbf{d \& f}: Group-level brain maps for seven fMRI sampling time points from the experiment block. Each group-level brain map at each time point is computed as an average over the relevance maps of each subject for this time point. Each group-level brain map is thresholded at the 90th percentile of its values. All brain maps were further projected onto the FsAverage5 surface template \cite{fischl2012freesurfer}. \textbf{g-h}: F1-score for each group-level brain map at each sampling time point of the experiment block. The F1-score quantifies the similarity between the group-level brain map and the results of the meta-analysis (c \& e) (for further details on the F1-score, see Section \ref{sec:results_group_level_maps_subject} and Supplementary Information Section 2). Red indicates the results of the F1-score comparison for the brain maps of DeepLight, whereas blue indicates the results of this comparison for the brain maps of the whole-brain lasso analysis (for further details on the F1-comparison for the whole-brain lasso analysis, see Section \ref{sec:results_ffa_ppa_block_anlaysis}).}
  \label{fig:DeepLight_vs_time}
\end{figure*}

To further probe DeepLight's ability to analyze single time points, we next studied the distribution of relevance values over the course of a single experiment block (Fig.\ \ref{fig:DeepLight_vs_time}). In particular, we plotted this distribution as a function of the fMRI sampling-time over all subjects for the first experiment block of the face and place stimulus classes in the second experiment run. We restricted this analysis to the face and place stimulus classes, as the neural networks involved in processing face and place stimuli, respectively, have been widely characterized (see for example \cite{haxby2001distributed} and \cite{heekeren2004general}). For a more detailed overview, we also created two videos for the two experiment blocks depicted in Fig.\ \ref{fig:DeepLight_vs_time} (Supplementary Videos 1 and 2). These videos display the temporal evolution of relevance values for each fMRI sample in the original fMRI sampling time. 

In the beginning of the experiment block, DeepLight was generally uncertain which cognitive state the observed brain samples belonged to, as it assigned similar probabilities to each of the cognitive states considered (Fig.\ \ref{fig:DeepLight_vs_time}\ a-b). As time progressed, however, DeepLight's certainty increased and it correctly identified the cognitive state underlying the fMRI samples. At the same time, it started assigning more relevance to the target ROIs of the face and place stimulus classes (Fig.\ \ref{fig:DeepLight_vs_time} c-f), as indicated by the increasing F1-scores resulting from a comparison of the brain maps at each sampling time point to the results of the meta-analysis (Fig.\ \ref{fig:DeepLight_vs_time}\ g-h; all brain maps were again thresholded at the 90th percentile for this comparison). Interestingly, the relevances started peaking in the target ROIs 5s after the onset of the experiment block. The temporal evolution of the relevances thereby mimics the hemodynamic response measured by the fMRI \cite{lindquist2009modeling}. 

To further evaluate the results of this analysis, we replicated it by the use of the whole-brain lasso group-level decoding model. In particular, we multiplied the fMRI samples of all test subjects collected at each sampling time point with the coefficient estimates of the whole-brain lasso group-level model (see Sections \ref{sec:wholebrain_lasso_details} and \ref{sec:classifier_training_details_group}). Subsequently, we averaged the resulting weighted fMRI samples within each sampling time point depicted in Fig.\ \ref{fig:DeepLight_vs_time}\ g-h and computed an F1-score for a comparison of the resulting average brain maps with the results of the meta-analysis (as described in section \ref{sec:results_decoding_comparison}). Interestingly, we found that the F1-scores of the whole-brain lasso analysis varied much less over the sequence of fMRI samples and were throughout lower than those of DeepLight. Thereby, indicating that the brain maps of the whole-brain lasso analysis exhibit comparably little variability over the course of an experiment block with respect to the target ROIs defined for the face and place stimulus classes.

\section{Discussion}
Neuroimaging data have a complex temporo-spatial dependency structure that renders modeling and decoding of experimental data a challenging endeavor. With DeepLight, we propose a new data-driven framework for the analysis and interpretation of whole-brain neuroimaging data that scales well to large datasets and is mathematically non-linear, while still maintaining interpretability of the data. To decode a cognitive state, DeepLight separates a whole-brain fMRI volume into its axial slices and processes the resulting sequence of brain slices by the use of a convolutional feature extractor and LSTM. Thereby, accounting for the spatially distributed patterns of whole-brain brain activity within and across axial slices. Subsequently, DeepLight relates cognitive state and brain activity, by decomposing its decoding decisions into the contributions of the single input voxels to these decisions with the LRP method. Thus, DeepLight is able to study the associations between brain activity and cognitive state on multiple levels of data granularity, from the level of the group down to the level of single subjects, trials and time points. 

To demonstrate the versatility of DeepLight, we have applied it to an openly available fMRI dataset of 100 subjects viewing images of body parts, faces, places and tools. With these data, we have shown that the DeepLight 1) decodes the underlying cognitive states more accurately from the fMRI data than conventional means of uni- and multivariate brain decoding, 2) improves its decoding performance better with growing datasets, 3) accurately identifies the physiologically appropriate associations between cognitive states and brain activity, 4) can study these associations on multiple levels of data granularity, from the level of the group down to the level of single subjects, trials and time points and 5) can capture the temporo-spatial variability of brain activity over sequences of single fMRI samples.

\subsection{Comparison to baseline methods}
\subsubsection{General linear model}
The GLM is conceptually different from the other neuroimaging analysis approaches considered in this work. It aims to identify an association between cognitive state and brain activity, by modeling (or predicting) the time series signal of a single voxel as a linear combination of a set of experiment predictors (see Section \ref{sec:GLM_details}). It is thereby limited in three meaningful ways that do not apply to DeepLight: First, the time series signal of a voxel is generally very noisy. The GLM treats each voxel's signal as independent of one another, thereby, not leveraging the evidence that is shared across the time series signal of multiple voxels. Second, even though the linear combination of a set of experiment predictors might be able to explain variance in the observed fMRI data, it does not necessarily provide evidence that this exact set of predictors is encoded in the neuronal response. Generally, the same linear model (in terms of its predictions) can be constructed from many different (even random) sets of predictors (for a detailed discussion of this "feature fallacy", see \cite{kriegeskorte2018interpreting}). The results of the GLM analysis thereby solely indicate that the measured neuronal response is highly structured and that this structure is preserved across individuals, whereas the labels assigned to its predictors might be arbitrary. Third, the performance of the GLM in predicting the response signal of a voxel is typically not evaluated on independent data, which leaves unanswered how well its results generalize to new data. 

\subsubsection{Searchlight analysis}
DeepLight generally outperformed the searchlight analysis in decoding the cognitive states from the fMRI data. In small datasets (here, the data of 10 or less subjects), however, the performance of the searchlight analysis was superior. In contrast to DeepLight, the searchlight analysis decodes a cognitive state from single clusters of only few voxels. Its input data, as well as the number of parameters in its decoding model, are thereby considerably smaller, leading to an overall lower risk of overfitting. Yet, this advantage comes at the cost of additional constraints that have to be considered when choosing between both approaches. If a cognitive state is associated with the activity of a small brain region only, the searchlight analysis will generally be more sensitive to the activity of this region than DeepLight, as it has learned a decoding model that is specific to the activity of the region. If, however, the cognitive state is not identifiable by the activity of a single brain region only, but solely in conjunction with the activity of another spatially distinct brain region, the searchlight analysis will not be able to identify this association, due to its narrow spatial focus. DeepLight, on the other hand, will generally be less sensitive to the specifics of the activity of a local brain region, but perform better in identifying a cognitive state from spatially wide-spread brain activity. When choosing between both approaches, one should therefore consider whether the assumed associations between brain activity and cognitive state specifically involve the activity of a local brain region only, or whether the cognitive state is associated with the activity of spatially distinct brain regions.  

\subsubsection{Whole-brain lasso}
In contrast to DeepLight, the whole-brain lasso analysis is based on a linear decoding model. It assigns a single coefficient weight to each voxel in the brain and makes a decoding decision by computing a weighted sum over the activity of an input fMRI volume. Importantly, due to the strong regularization that is applied to the coefficients during the training, many coefficients equal 0. The resulting set of coefficients thereby resembles a brain mask, defining a set of fixed brain regions whose activity the whole-brain lasso utilizes to decode a cognitive state. DeepLight, on the other hand, utilizes a hierarchical structure of non-linear transforms of the fMRI data. It projects each fMRI volume into a more abstracted, higher-level space. This abstracted (and more flexible) view enables DeepLight to better account for the variable patterns of brain activity underlying a cognitive state (within and across individuals). This ability is exemplified in Fig.\ \ref{fig:DeepLight_vs_time}, as well as Supplementary Videos 1 - 2, where we visualize the variable patterns of brain activity that DeepLight associates with the face and place stimulus classes throughout an experiment block. The relevance patterns of DeepLight mimic the hemodynamic response and peak in the ROIs 5-10s after the onset of the experiment block. Importantly, we find that the whole-brain lasso does not exhibit such temporo-spatial variability.

\subsection{Disentangling temporally distinct associations between cognitive state and brain activity}
DeepLight's ability to identify a cognitive state through variable patterns of brain activity makes it ideally suited for the analysis of the fine-grained temporo-spatial distribution of brain activity over temporal sequences of fMRI samples. For example, Hunt and Hayden \cite{hunt2017distributed} recently raised the question whether the neural networks involved in reward-based decision making can be subdivided into a set of spatially distinct and temporally discrete network components, or whether the underlying networks act in parallel, with highly recurrent activity patterns. Answering this question is difficult with conventional approaches to the analysis of neuroimaging data, such as the baseline methods included in this paper. These often learn a fixed mapping between brain activity and cognitive state, by aggregating over the information provided by many fMRI samples (e.g., by estimating a single coefficient weight for each voxel from sequences of fMRI data). The resulting brain maps thereby only indicate whether there exist spatially distinct brain regions that are associated with a cognitive state, without providing any insight whether the activity patterns are temporally discrete. While these methods can be adapted to specifically account for the temporal differences in the activity patterns of these regions (e.g., by analyzing different time points independent of one another), these adaptations often require specific hypotheses about the studied temporal differences (e.g., by needing to specify the different time points to analyze). DeepLight, on the other hand, operates purely data-driven and is thereby able to autonomously identify an association between cognitive state and spatially distinct patterns of brain activity at temporally discrete time points.

\subsection{Integrative analysis of multimodal neuroimaging data}
The ability of DeepLight to analyze the temporo-spatial distribution of brain activity might, at first sight, seem excessive for fMRI data, which is known to suffer from a low temporal resolution, despite its high spatial accuracy. Yet, DeepLight is not bound to fMRI data, but can be easily extended to other neuroimaging modalities. One such complementary modality, with a higher temporal, but lower spatial resolution, is Electroencephalography (EEG). While a plethora of analysis approaches have been proposed for the integrative analysis of EEG and fMRI data, these often incorporate restrictive assumptions to enable the integrative statistical analysis of these two data types, with clearly distinct spatial, temporal and physiological properties (for a detailed review, see \cite{jorge2014eeg}). DeepLight, on the other hand, provides a  data-driven analysis framework. By providing both, EEG and fMRI data as separate inputs to the DL model, DeepLight could learn the fine-grained temporal structure of brain activity from the EEG data, while utilizing the fMRI data to localize the spatial brain regions underlying this activity. Recently, researchers have already demonstrated the usefulness of interpretable DL methods for the analysis of EEG data \cite{sturm2016interpretable}.

\subsection{Extending DeepLight}
Lastly, we would like to highlight several possible extensions of the DeepLight approach, resulting from its flexible and modular architecture. First, DeepLight can be extended to specifically account for the temporo-spatial distribution of brain activity over sequences of fMRI samples, by the addition of another recurrent network layer. This layer would process each of the higher-level whole-brain representations resulting from the currently proposed architecture. This extension would enable DeepLight to more specifically account for the time points at which the activity of a brain region is most informative about a cognitive state. Second, DeepLight can be extended to the integrative analysis of neuroimaging data from different cognitive tasks and experiments. For example, by adding one neuron to the output layer for each cognitive state from each task. This extension would enable a more thorough analysis of the differences (and similarities) between the associations of cognitive state and brain activity across multiple task domains.

\section*{Acknowledgements}
This work was supported by the German Federal Ministry for Education and Research through the Berlin Big Data Centre (01IS14013A), the Berlin Center for Machine Learning (01IS18037I) and the TraMeExCo project (01IS18056A). Partial funding by the German Research Foundation (DFG) is acknowledged (EXC 2046/1, project-ID: 390685689). KRM is also supported by the Information \& Communications Technology Planning \& Evaluation (IITP) grant funded by the Korea government (No. 2017-0-00451).

\bibliography{DeepLight_references}{}
\bibliographystyle{apalike}

\newpage
\section*{Supplementary Information}

\renewcommand{\thefigure}{S\arabic{figure}}\setcounter{figure}{0} 
\renewcommand{\thetable}{S\arabic{table}}\setcounter{table}{0}

\subsection*{1. NeuroSynth}
\label{sec:NeuroSynth_details}
The goal of NeuroSynth \cite{yarkoni2011large} is to provide an automated meta-analysis database relating cognitive states and brain activity. For specific cognitive states (i.e., ``pain''), the NeuroSynth database incorporates a large record of neuroimaging studies that used this term at a high frequency (>1 in 1000 words) in the article text. For these studies, the database includes the activation coordinates from all tables that are reported in these studies, producing a large set of {\em term-to-activation} mappings. Based on these mappings, NeuroSynth provides two types of tests: a {\em uniformity test}, indicating whether the probability that an article reports a specific brain activation is different, if it includes a specific term, compared to when brain activation would be distributed uniformly throughout gray matter and an {\em association test}, indicating whether the probability that a research article reports a specific brain activation is different, if it includes a specific term, compared to when it does not.

For our analyses, we used the latter, association test, as recommended by the NeuroSynth authors \cite{yarkoni2011large}, and extracted the thresholded ($p \leq 0.01$, FDR corrected) brain maps for the four stimulus classes (indicated by the terms "body'', "face'', "place'' and "tools''). These maps indicate a Z-value for the previously described association test at each coordinate in the MNI-space.

\subsection*{2. F1-score}
\label{sec:f1_score_details}
The F1-score for a binary classifier and a given dataset is defined as the harmonic mean of its precision and recall:

\begin{equation}
    F_{1} = 2 \frac{precision \cdot recall}{precision + recall}
\end{equation}

Here, the classifier's precision is defined as the fraction of samples in the dataset that it correctly classified as positives, given the total number of samples that it classified as positive in the dataset, whereas its recall describes the fraction of samples that it correctly classified as positive, given the overall number of positive samples in the dataset.

\clearpage
\begin{figure}
  \centering
  \includegraphics[width=0.75\linewidth]{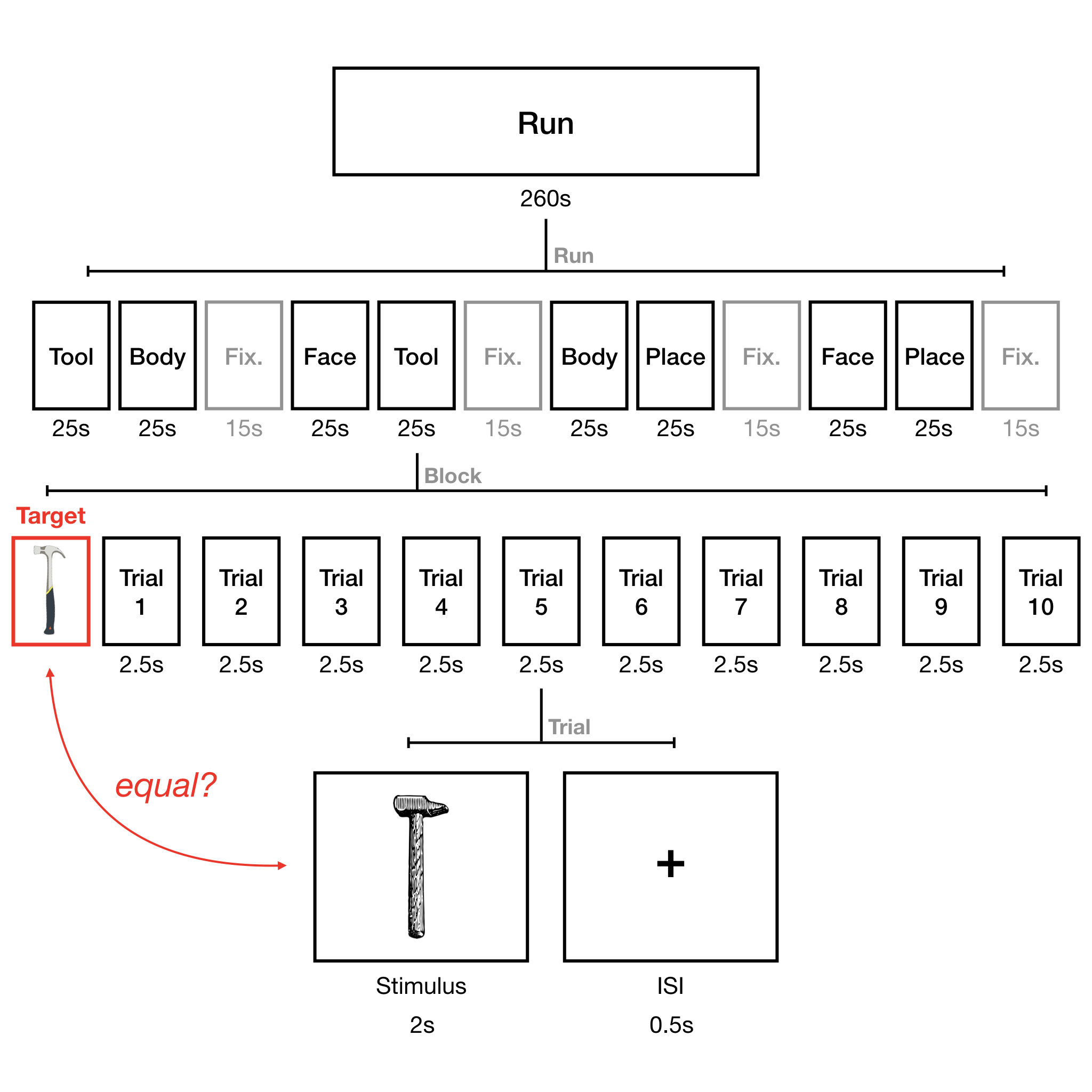}
  \caption{Experimental Paradigm. 100 subjects completed two experiment runs in the fMRI. Each run consisted of eight task and four fixation blocks. The four stimulus types (body, face, place, tool) were presented in separate blocks. Each task block consisted of 10 trials. In each trial, a stimulus was presented for 2s, followed by a 500ms interstimulus interval. Subjects performed an N-back task, in which they were asked to respond "target" when the currently presented stimulus was the same as a target stimulus. The target was either presented at the beginning of the block (0-back) or subjects were asked to indicate whether the current stimulus was the same as the stimulus two back (2-back). Half of the blocks used a 2-back and the other half a 0-back condition. For illustrative purposes only the 0-back task is depicted.}
  \label{fig:task}
\end{figure}

\clearpage
\begin{figure}
	\centering
 	\includegraphics[width=0.75\linewidth]{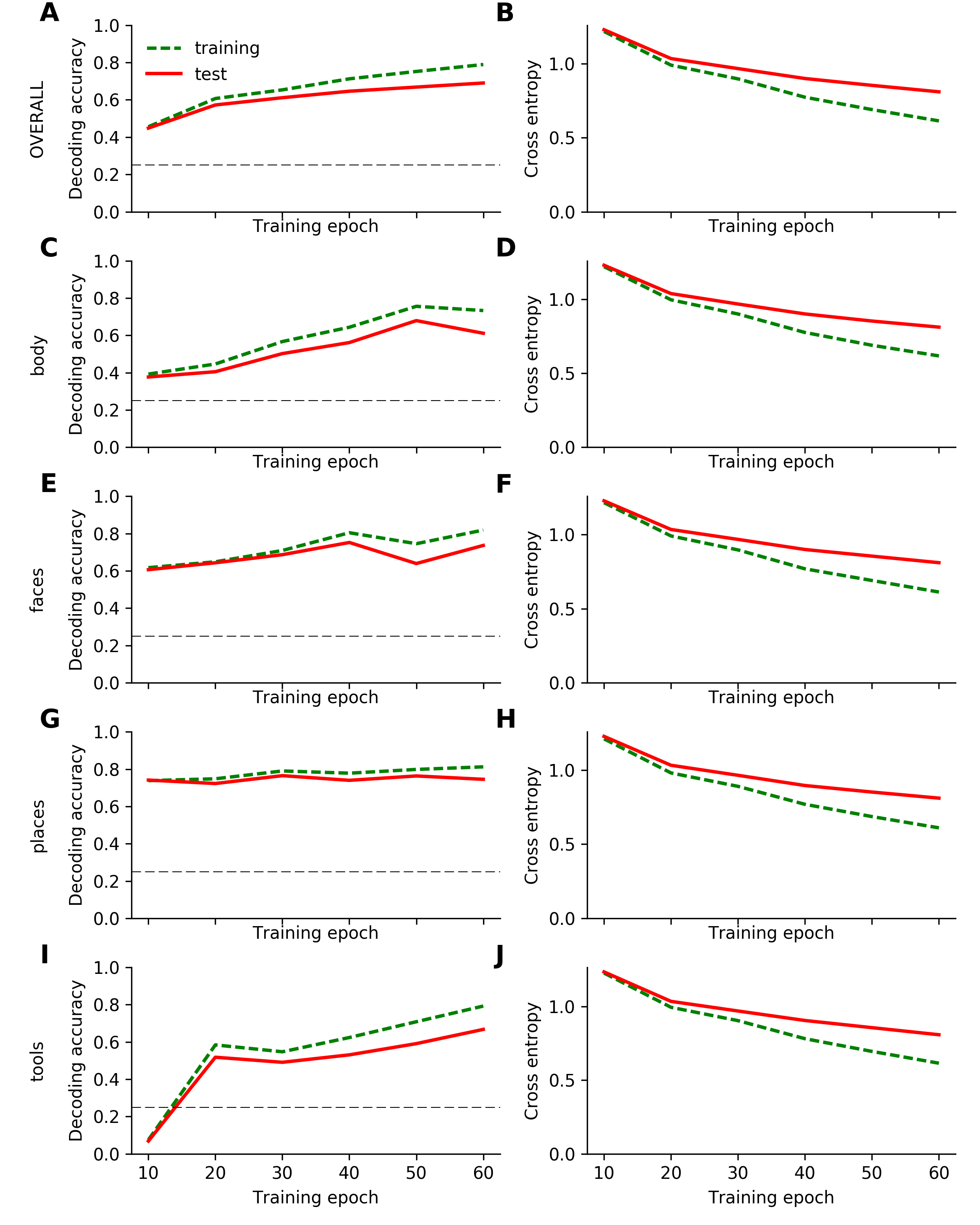}
 	\caption{DeepLight's training statistics as a function of the training epochs. Red indicates the test data, whereas green indicates the training dataset. An epoch is defined as a full iteration over the training data. We define decoding accuracy as the fraction of samples that were classified correctly. For further details on DeepLight's training, see the Methods Section of the main text.}
 	\label{fig:training_statistics}
\end{figure}

\clearpage
\begin{figure}
	\centering
 	\includegraphics[width=\linewidth]{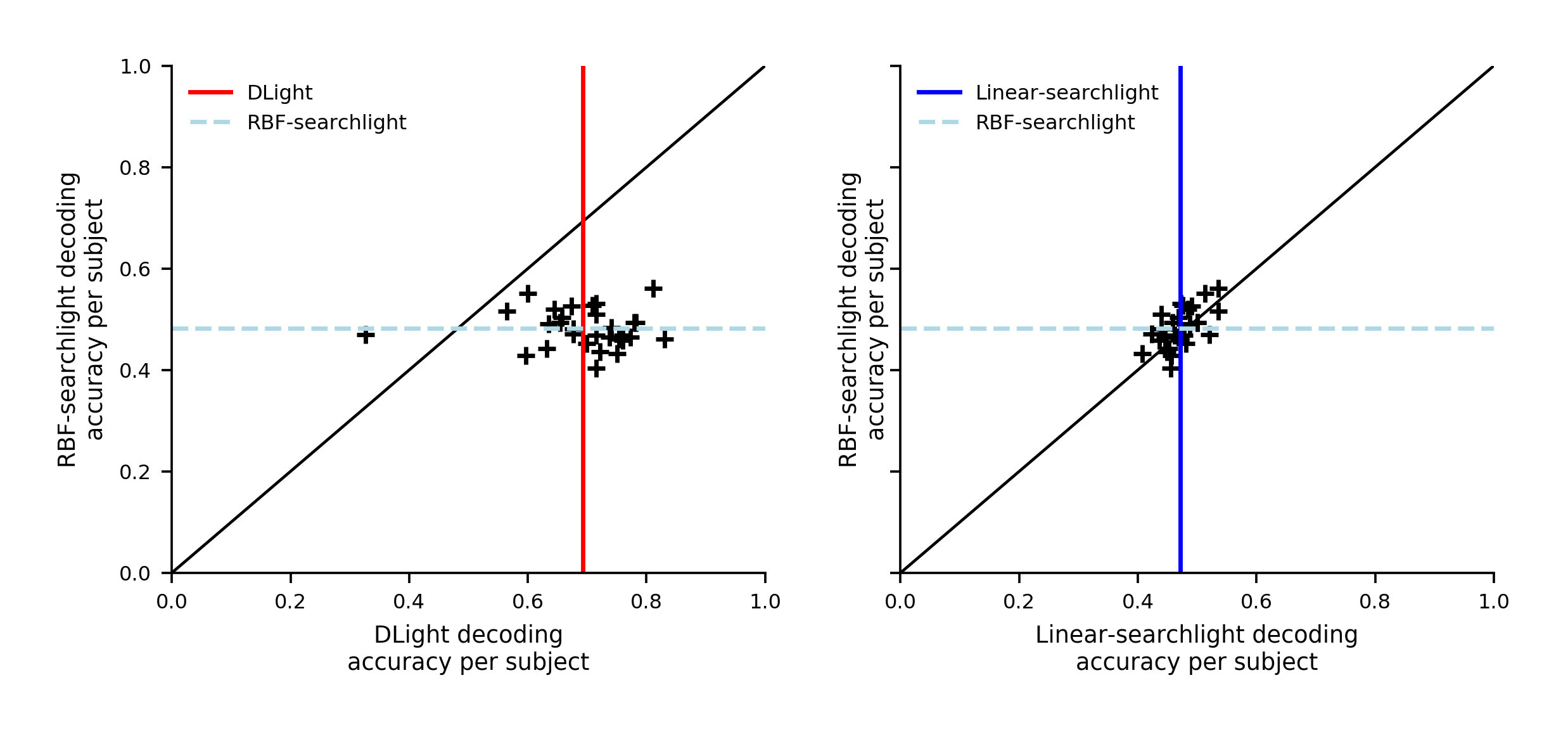}
 	\caption{Out of sample decoding performance comparison of the searchlight analysis with a linear-kernel SVM and a non-linear radial basis function (RBF) kernel SVM. We trained each searchlight on the data of the first experiment run of a subject, before predicting the cognitive state for each volume of the second experiment run (for details on the estimation procedures, see the Methods Section of the main text). We performed this prediction exercise only within the data of the subjects in the held-out test dataset. We fixed the searchlight radius to 5.6mm, while we set $\gamma$ parameter of the RBF-kernel to 1 across all subjects. {\it Left}: Decoding performance comparison of the RBF-kernel SVM with DeepLight. {\it Right}: Decoding performance comparison of the RBF-kernel SVM with the linear-kernel SVM. Each cross represents an individual subject. Colored lines indicate averages across subjects. For an overview of the statistical results of the comparison, see the Results Section of the main text.}
 	\label{fig:rbf_searchlight}
\end{figure}

\clearpage
\begin{figure}
	\centering
 	\includegraphics[width=\linewidth]{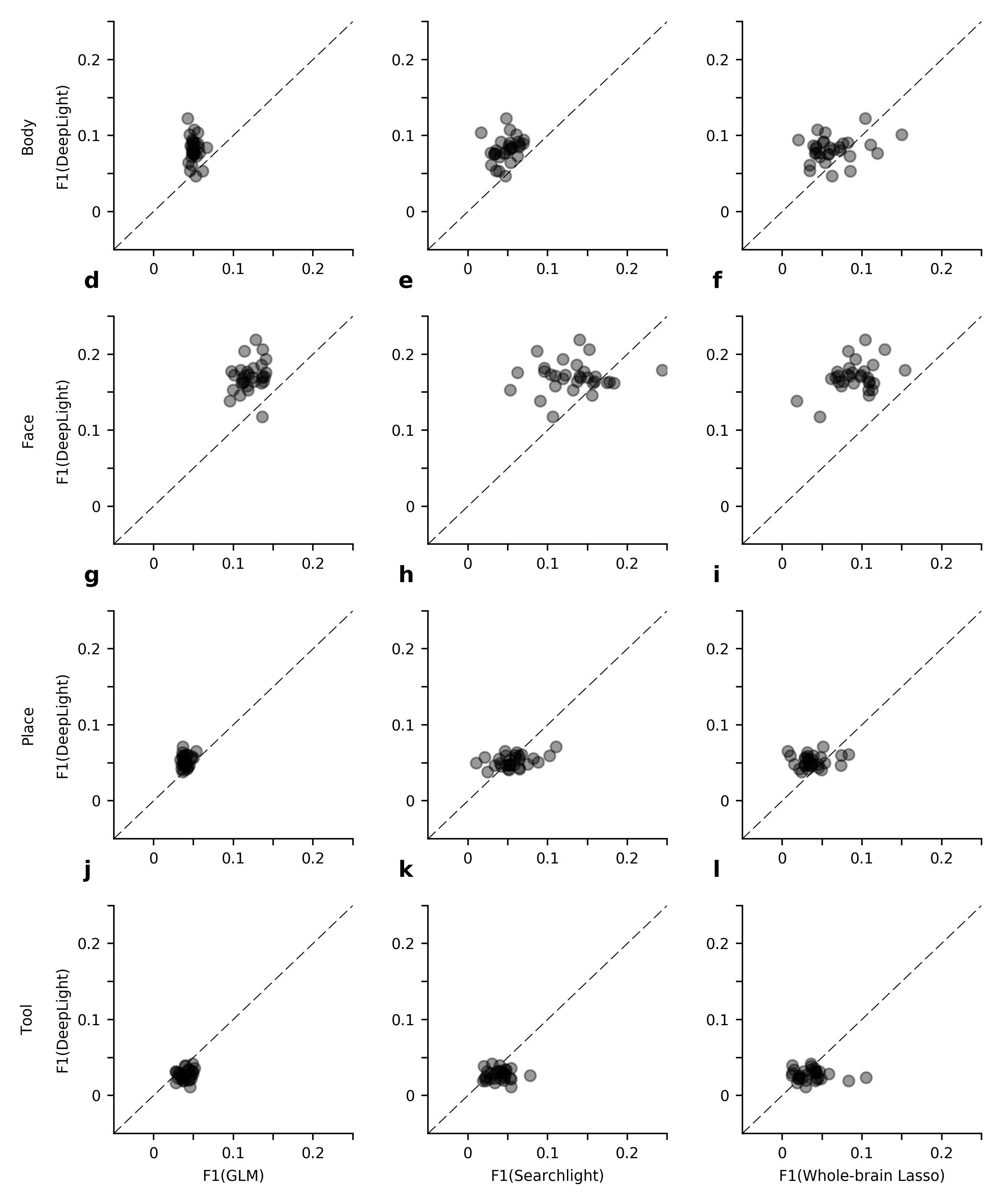}
 	\caption{Comparison of the subject-level F1-scores for the brain maps of DeepLight, the GLM, searchlight analysis and whole-brain lasso. The brain maps of DeepLight, the searchlight analysis and whole-brain lasso were thresholded at the 90th percentile of the values within each map, whereas the GLM brain maps were thresholded at a P-value of 0.005 (uncorrected). Points indicate individual subjects. For an overview of the statistical results of the comparison, see the Results Section of the main text.}
 	\label{fig:f1_comparison_subject_level}
\end{figure}

\clearpage
\begin{table}[htp]
\centering
\begin{tabular}{
  l
  S[table-format=2.2]
  S[table-format=2.0]
}
\toprule
\textrm{Test Subject} & {$C$} & \textrm{Decoding Accuracy}\\
\midrule
1 & 18.74 & 33\% \\
2 & 0.26 & 33\% \\
3 & 0.09 & 42\% \\
4 & 0.66 & 42\% \\
5 & 0.22 & 47\% \\
6 & 0.26 & 49\% \\
7 & 0.79 & 37\% \\
8 & 22.57 & 42\% \\
9 & 12.92 & 26\% \\
10 & 83.02 & 40\% \\
11 & 39.44 & 27\% \\
12 & 22.57 & 37\% \\
13 & 39.44 & 32\% \\
14 & 0.26 & 42\% \\
15 & 0.79 & 38\% \\
16 & 3.51 & 38\% \\
17 & 3.51 & 41\% \\
18 & 0.26 & 22\% \\
19 & 0.15 & 36\% \\
20 & 27.19 & 38\% \\
21 & 32.75 & 45\% \\
22 & 7.39 & 34\% \\
23 & 6.14 & 41\% \\
24 & 7.39 & 45\% \\
25 & 0.66 & 29\% \\
26 & 4.23 & 38\% \\
27 & 2.92 & 33\% \\
28 & 47.51 & 42\% \\
29 & 12.92 & 36\% \\
30 & 1.67 & 23\% \\
\midrule[\heavyrulewidth]
\end{tabular}
\caption{Selected regularization strength parameters for the subject-level whole-brain lasso analysis (for details on the underlying subject-level grid search procedure, see the Methods Section of the main text). For each subject, the selected regularization parameter ($C$) and resulting decoding accuracy are presented.}
\end{table}

\clearpage
\begin{table}[htp]
\centering
\begin{tabular}{
  l
  S[table-format=2.2]
}
\toprule
{$\lambda$} & \textrm{Decoding Accuracy}\\
\midrule
1e-7 & 46.62\% \\
1e-6 & 46.40\% \\
1e-5 & 47.82\% \\
{\textbf{0.0001}} & {\textbf{48.11\%}} \\
0.0002 & 47.24\% \\
0.0003 & 47.42\% \\
0.0005 & 46.11\% \\
0.0008 & 47.03\% \\
0.001 & 45.53\% \\
0.002 & 45.58\% \\
0.004 & 44.98\% \\
0.007 & 44.86\% \\
0.01 & 42.89\% \\
0.02 & 40.69\% \\
0.03 & 37.14\% \\
0.06 & 35.48\% \\
0.1 & 26.57\% \\
0.2 & 25.13\% \\
0.3 & 25.00\% \\
0.5 & 25.09\% \\
\midrule[\heavyrulewidth]
\end{tabular}
\caption{$\lambda$ parameters of the group-level whole-brain lasso anlaysis that were evaluated in the grid search procedure (for details on the group-level grid search, see the Methods Section of the main text). The respective decoding accuracy in the test dataset is given for each $\lambda$ value.}
\end{table}

\clearpage
\begin{table}[htp]
\centering
\addtolength{\tabcolsep}{5pt} 
\begin{tabular}{
  S[table-format=1.5]
  S[table-format=2.0]
  S[table-format=2.0]
  S[table-format=2.0]
  S[table-format=2.0]
  S[table-format=2.0]
}
\toprule
\textrm{P} & \textrm{Percentile} & \textrm{N(DeepLight)} & \textrm{N(GLM)} & \textrm{N(Searchlight)} & \textrm{N(whole-brain lasso)}\\
\midrule
0.05 & 85 & {\textbf{20}} & 0 & 0 & 10\\
0.05 & 90 & {\textbf{24}} & 0 & 0 & 6\\
0.05 & 95 & {\textbf{26}} & 0 & 0 & 4\\

0.005 & 85 & {\textbf{20}} & 0 & 0 & 10\\
0.005 & 90 & {\textbf{24}} & 0 & 0 & 6\\
0.005 & 95 & {\textbf{26}} & 0 & 0 & 4\\

0.0005 & 85 & {\textbf{20}} & 0 & 0 & 10\\
0.0005 & 90 & {\textbf{24}} & 0 & 0 & 6\\
0.0005 & 95 & {\textbf{26}} & 0 & 0 & 4\\

0.00005 & 85 & {\textbf{20}} & 0 & 0 & 10\\
0.00005 & 90 & {\textbf{24}} & 0 & 0 & 6\\
0.00005 & 95 & {\textbf{26}} & 0 & 0 & 4\\
\midrule[\heavyrulewidth]
\end{tabular}
\caption{Number of subjects from the held-out test dataset with the highest subject-level F1-score for body stimuli for each analysis approach and combination of percentile- and P-threshold (for details on the F1-score comparison, see the Results Section of the main text and Supplementary Information Section 2).}
\end{table}

\clearpage
\begin{table}[htp]
\centering
\addtolength{\tabcolsep}{5pt} 
\begin{tabular}{
  S[table-format=1.5]
  S[table-format=2.0]
  S[table-format=2.0]
  S[table-format=2.0]
  S[table-format=2.0]
  S[table-format=2.0]
}
\toprule
\textrm{P} & \textrm{Percentile} & \textrm{N(DeepLight)} & \textrm{N(GLM)} & \textrm{N(Searchlight)} & \textrm{N(whole-brain lasso)}\\
\midrule
0.05 & 85 & {\textbf{25}} & 1 & 4 & 0\\
0.05 & 90 & {\textbf{24}} & 1 & 5 & 0\\
0.05 & 95 & {\textbf{20}} & 3 & 7 & 0\\

0.005 & 85 & {\textbf{25}} & 1 & 4 & 0\\
0.005 & 90 & {\textbf{24}} & 1 & 5 & 0\\
0.005 & 95 & {\textbf{20}} & 3 & 7 & 4\\

0.0005 & 85 & {\textbf{25}} & 1 & 4 & 0\\
0.0005 & 90 & {\textbf{24}} & 1 & 5 & 0\\
0.0005 & 95 & {\textbf{20}} & 3 & 7 & 0\\

0.00005 & 85 & {\textbf{25}} & 1 & 4 & 0\\
0.00005 & 90 & {\textbf{24}} & 1 & 5 & 0\\
0.00005 & 95 & {\textbf{20}} & 3 & 7 & 0\\
\midrule[\heavyrulewidth]
\end{tabular}
\caption{Number of subjects from the held-out test dataset with the highest subject-level F1-score for face stimuli for each analysis approach and combination of percentile- and P-threshold (for details on the F1-score comparison, see the Results Section of the main text and Supplementary Information Section 2).}
\end{table}

\clearpage
\begin{table}[htp]
\centering
\addtolength{\tabcolsep}{5pt} 
\begin{tabular}{
  S[table-format=1.5]
  S[table-format=2.0]
  S[table-format=2.0]
  S[table-format=2.0]
  S[table-format=2.0]
  S[table-format=2.0]
}
\toprule
\textrm{P} & \textrm{Percentile} & \textrm{N(DeepLight)} & \textrm{N(GLM)} & \textrm{N(Searchlight)} & \textrm{N(whole-brain lasso)}\\
\midrule
0.05 & 85 & 5 & 3 & {\textbf{17}} & 5\\
0.05 & 90 & 10 & 0 & {\textbf{16}} & 4\\
0.05 & 95 & {\textbf{14}} & 0 & 13 & 3\\

0.005 & 85 & 5 & 3 & {\textbf{17}} & 5\\
0.005 & 90 & 10 & 0 & {\textbf{16}} & 4\\
0.005 & 95 & {\textbf{14}} & 0 & 13 & 3\\

0.0005 & 85 & 5 & 3 & {\textbf{17}} & 5\\
0.0005 & 90 & 10 & 0 & {\textbf{16}} & 4\\
0.0005 & 95 & {\textbf{14}} & 0 & 13 & 3\\

0.00005 & 85 & 5 & 3 & {\textbf{17}} & 5\\
0.00005 & 90 & 10 & 0 & {\textbf{16}} & 4\\
0.00005 & 95 & {\textbf{14}} & 0 & 13 & 3\\
\midrule[\heavyrulewidth]
\end{tabular}
\caption{Number of subjects from the held-out test dataset with the highest subject-level F1-score for place stimuli for each analysis approach and combination of percentile- and P-threshold (for details on the F1-score comparison, see the Results Section of the main text and Supplementary Information Section 2).}
\end{table}

\clearpage
\begin{table}[htp]
\centering
\addtolength{\tabcolsep}{6pt} 
\begin{tabular}{
  S[table-format=1.5]
  S[table-format=2.0]
  S[table-format=2.0]
  S[table-format=2.0]
  S[table-format=2.0]
  S[table-format=2.0]
}
\toprule
\textrm{P} & \textrm{Percentile} & \textrm{N(DeepLight)} & \textrm{N(GLM)} & \textrm{N(Searchlight)} & \textrm{N(whole-brain lasso)}\\
\midrule
0.05 & 85 & 1 & 10 & {\textbf{12}} & 7\\
0.05 & 90 & 1 & 9 & {\textbf{13}} & 7\\
0.05 & 95 & 2 & {\textbf{10}} & 9 & 9\\

0.005 & 85 & 1 & 10 & {\textbf{12}} & 7\\
0.005 & 90 & 1 & 9 & {\textbf{13}} & 7\\
0.005 & 95 & 2 & {\textbf{10}} & 9 & 9\\

0.0005 & 85 & 1 & 10 & {\textbf{12}} & 7\\
0.0005 & 90 & 1 & 9 & {\textbf{13}} & 7\\
0.0005 & 95 & 2 & {\textbf{10}} & 9 & 9\\

0.00005 & 85 & 1 & 10 & {\textbf{12}} & 7\\
0.00005 & 90 & 1 & 9 & {\textbf{13}} & 7\\
0.00005 & 95 & 2 & {\textbf{10}} & 9 & 9\\
\midrule[\heavyrulewidth]
\end{tabular}
\caption{Number of subjects from the held-out test dataset with the highest subject-level F1-score for tool stimuli for each analysis approach and combination of percentile- and P-threshold (for details on the F1-score comparison, see the Results Section of the main text and Supplementary Information Section 2).}
\end{table}

\end{document}